%% file: neurips_2026.tex
\documentclass{article}
\PassOptionsToPackage{numbers, compress}{natbib}
\usepackage[preprint]{neurips_2026}

\usepackage[utf8]{inputenc} 
\usepackage[T1]{fontenc}    
\usepackage{hyperref}       
\usepackage{url}            
\usepackage{booktabs}       
\usepackage{amsfonts}       
\usepackage{nicefrac}       
\usepackage{microtype}      
\usepackage{xcolor}         

\usepackage[utf8]{inputenc}
\usepackage{amsfonts}
\usepackage{amssymb}
\usepackage{amsmath}
\usepackage{natbib}
\usepackage{hyperref}
\usepackage{comment}
\usepackage{bm}
\usepackage{enumitem}
\usepackage{tikz}
\usepackage{subcaption}
\usepackage{wrapfig}
\usepackage{rotating}
\usetikzlibrary{arrows.meta,positioning,calc}

\usepackage{amsthm}
\newtheorem{theorem}{Theorem}[section]
\newtheorem{corollary}{Corollary}[theorem]

\newtheorem{definition}{Definition}[section]
\newtheorem{proposition}{Proposition}[section]

\usepackage{centernot}

\DeclareMathOperator{\indep}{\perp\!\!\!\perp}

\usetikzlibrary{positioning,arrows.meta}

\title{Probabilities of Causation in\\Nondeterministic Causal Models}

\author{%
  David S. Watson\\
  Google DeepMind\\
  King's College London\\
  \texttt{dswatson@google.com}\\
  \And
  Sander Beckers\\
  University College London\\
  \texttt{srekcebrednas@gmail.com}
}

\begin{document}

\maketitle

\begin{abstract}
    We study probabilities of causation in nondeterministic causal models, where mechanisms may remain intrinsically stochastic even after all latent variables are fixed. Focusing on binary treatments and outcomes, we develop a generic identification program that represents observational and interventional data as moment constraints on latent stochastic mechanism types and computes minimal/maximal values of counterfactual quantities via constrained optimization. 
    Within this framework, we derive sharp analytic bounds on the probabilities of causation under several assumptions tailored to nondeterministic models.
    Resulting identification regions are often more informative than those attainable under deterministic alternatives. We illustrate the practical relevance of these results in a case study on personalized treatment decisions and experiments on semi-synthetic datasets.
    Our results clarify how classical bounds relate to nondeterministic causal semantics, generalizing previous works and providing a unified route from qualitative structural restrictions to computable partial identification regions.
\end{abstract}

 \input{main_CLeaR}

\begin{ack}
Thanks to Ricardo Silva for his helpful feedback.
\end{ack}

\bibliographystyle{plainnat}
\small{
\bibliography{biblio} 
}


\appendix

\input{appx_CLeaR}

\end{document}

%% file: main_CLeaR.tex
\section{Introduction}
Suppose that variable $X$ (say, smoking) causes variable $Y$ (say, cancer).
How should we evaluate the strength of this relationship? 
One influential proposal relies on counterfactual quantities known as the \textit{probabilities of causation} \citep{tian_pearl_2000}. 
These answer questions not just about whether smoking and cancer co-occur, or whether randomly assigning people to a smoking group increases their chance of cancer.
Probabilities of causation are strictly more informative---telling us, for instance, how likely someone would be to develop cancer \textit{had} they smoked, given that they did not smoke and do not have cancer. 
Such counterfactual reasoning represents the topmost rung of Pearl's ladder of causation \citep{pearl2009causality, bareinboim2022}. 
These sorts of inferences are essential for mechanistic understanding and scientific explanation \citep{woodward2003making, halpern_pearl}, and have become increasingly prevalent in explainable AI \citep{karimi2021, beckers2022, watson2022, verma2024}. 

Formulating a rigorous logic of counterfactuals has proven contentious.
Classical philosophical accounts, most notably that of \citet{lewis1973}, evaluate counterfactuals by comparing the actual world to suitably ``close'' possible worlds, but leave substantial work to the choice of similarity relation.
Modern causal inference instead typically formalizes counterfactuals through potential outcomes \citep{rubin1974} or structural causal models \citep{pearl2009causality}, where interventions modify a system of structural equations.
This approach has itself attracted criticism. \citet{dawid2000}, for example, questions the need for counterfactual variables in many causal inference problems and emphasizes the assumptions required to interpret unobserved potential responses.
More recently, several authors have sought to define counterfactual semantics directly for probabilistic causal mechanisms.
\citet{galhotra_halpern_2024} develop such a semantics for causal Bayesian networks, while \citet{beckers2025nondeterministic} relaxes the usual deterministic uniqueness assumptions altogether, yielding probabilistic nondeterministic structural causal models (PNSCMs), as opposed to the standard Pearlian probabilistic deterministic structural causal models (PDSCMs).

PNSCMs have the advantage that they do not require latent variables in order to explain the presence of a probabilistic relationship. In Markovian graphs, the absence of latent variables makes counterfactual identification possible. However, this advantage disappears under latent confounding, which poses nontrivial challenges on any causal semantics. A common strategy in deterministic models is to introduce assumptions such as exogeneity and monotonicity to tighten the range of feasible solutions for counterfactuals, even providing point identification in some cases. 
Such assumptions can be ill-suited to PNSCMs, since they require extreme probabilities given a complete assignment to a variable's causal parents.

We study the problem of identifying probabilities of causation when mechanisms are irreducibly stochastic. 
Specifically, we examine the semi-Markovian case, where latent confounding makes point identification generally impossible. 
Our main contributions include: (1) a formal statement of the constrained optimization task that presents causal probability bounds as the solution to a generalized nonlinear program; (2) provably sharp analytic bounds that apply under a range of structural assumptions well-suited to nondeterministic models; (3) a case study on personalized treatment decisions; and (4) experimental results on semi-synthetic datasets that confirm the utility of our approach.


\section{Background}\label{sect:background}
\defcitealias{tian_pearl_2000}{TP}

\begin{wrapfigure}{r}{0.33\textwidth}
  \centering
  \vspace{-\intextsep}
  \begin{tikzpicture}[
      >={Stealth[length=2mm,width=1.6mm]},
      obs/.style={circle,draw,minimum size=7mm,inner sep=0pt},
      lat/.style={circle,draw,dashed,minimum size=7mm,inner sep=0pt},
      arr/.style={->,thick}
    ]
    \node[obs] (X) at (0,0)   {$X$};
    \node[obs] (Y) at (2.4,0) {$Y$};
    \node[lat] (U) at (1.2,1.3) {$U$};
    \draw[arr] (X) -- (Y);
    \draw[arr] (U) -- (X);
    \draw[arr] (U) -- (Y);
  \end{tikzpicture}
  \caption{DAG for the confounded model $\mathcal{M}$. The dashed node is unobserved.}
  \label{fig:dag1}
\end{wrapfigure}
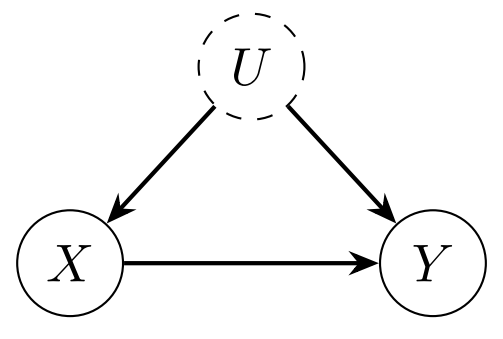

Let $X \in \{x, x'\}$ be a direct cause of $Y \in \{y, y'\}$ in model $\mathcal M$, where we encode these values as $x \equiv 1$, $x' \equiv 0$ and $y \equiv 1$, $y' \equiv 0$ whenever convenient. 
Let $U \in \mathcal U$ be a latent variable that is not caused by $X$ or $Y$, but which may be a cause of one or both. In the latter case, we say that $U$ is a \textit{latent confounder}. 
The resulting directed acyclic graph (DAG) is presented in Fig.~\ref{fig:dag1}. 
We will refer to observational probabilities such as $p(y \mid x)$ as $\mathcal L_1$ data, to indicate the first rung of \citet{pearl2009causality}'s ladder of causation. 
We write $p(y_x)$ to denote the probability that $Y=y$ under a hard intervention of the form $X \gets x$, which assigns value $x$ to variable $X$ no matter the latent state $u$. 
This is an $\mathcal L_2$ quantity, sometimes but not always identifiable from $\mathcal L_1$ data \citep{shpitser2008}. Our focus in this work is \textit{counterfactuals}, which correspond to the highest rung of Pearl's ladder, $\mathcal L_3$. 
Such quantities are encoded in the full model $\mathcal M$, but are underdetermined by the data $\mathcal D := \left\{p(X, Y), p(Y_x), p(Y_{x'})\right\}$.
We will generally assume complete knowledge of $\mathcal L_1$ and $\mathcal L_2$ distributions, as well as strict positivity of all quantities in $\mathcal D$ (including relevant marginals).\footnote{Though not strictly necessary, positivity avoids certain degenerate behavior when dividing by zero, and can be easily enforced via Laplacian smoothing.}
Let $\mathfrak M$ denote the class of admissible models, i.e. fully specified causal models on binary $X, Y$ consistent with the structure $X \rightarrow Y$. 
We write $\mathfrak M(\mathcal D) := \{\mathcal M \in \mathfrak M: \mathcal M \models \mathcal D\}$ to denote the set of admissible models compatible with $\mathcal D$. 

In a seminal article, \citet{tian_pearl_2000} (henceforth \citetalias{tian_pearl_2000}) define the probability of necessity (PN), the probability of sufficiency (PS), and the probability of necessity and sufficiency (PNS):
\begin{align*}
    \mathrm{PN} := p(y'_{x'} \mid x, y), \quad 
    \mathrm{PS} := p(y_x \mid x', y'), \quad 
    \mathrm{PNS} := p(y_x, y'_{x'}).
\end{align*}
To continue with our running example, PN answers the question: if I smoke and have cancer, how likely would I be to not have cancer had I not smoked? 
Inversely, PS answers the question: if I do not smoke and do not have cancer, how likely would I be to have cancer had I smoked? 
PNS represents the joint probability of both potential outcomes. 
Collectively, these three quantities constitute the probabilities of causation. Though these $\mathcal L_3$ estimands are not generally point identifiable from $\mathcal L_1$ and $\mathcal L_2$ data, \citetalias{tian_pearl_2000} derive a linear program (LP) for computing sharp bounds on each under a range of different assumptions, including exogeneity and monotonicity (formally defined below).

PNSCMs replace deterministic structural equations with conditional probability kernels, allowing stochastic outcomes even given all causal parents \citep{beckers2025nondeterministic}.
Their counterfactual semantics employ \emph{actualized refinement}: observed parent--child assignments are preserved whenever the same parental configuration obtains (thereby ensuring consistency), while alternative configurations retain their stochastic mechanisms. 
Joint counterfactual probabilities are then defined by a product coupling conditional on the actual world, followed by marginalization over worlds.

We refer the reader to \citep{beckers2025b} for a detailed justification of this semantics and here only sketch the basic idea. PNSCMs are meant to represent models that are {\em causally complete}, which means that there are no missing parents of endogenous variables. (This condition is thus stronger than causal sufficiency.) For our model $\mathcal M$, this means that $U$ is assumed to be the only latent parent of $Y$. Conditional on all its parents, any remaining variation in a child variable must be intrinsic, and therefore the counterfactual mechanism consists of the same conditional distributions as the actual one. 
Since the actual mechanism assigns a unique value -- namely the actual one -- to the child variable conditional on its actual parents, so does the counterfactual mechanism.
For $\mathcal M$, it follows that $X$, $Y_x$ and $Y_{x'}$ are mutually independent conditional on $U=u$, with $p(y_x \mid u) = p(y \mid x, u)$ and $p(y_{x'} \mid u) = p(y \mid x', u)$ (see Appx.~\ref{appx:linear} for the derivation).
With no latent variables, this yields
$\mathrm{PN}=p(y'\mid x')$,
$\mathrm{PS}=p(y\mid x)$, and
$\mathrm{PNS}=p(y\mid x)~p(y'\mid x')$.
However, these point-identification results do not extend to models with latent confounding.

An important preliminary result delimits what nondeterminism does and does not buy us.
\begin{proposition}[Inheritance]
\label{thm:inheritance}
The sets of joint distributions over $(X,Y_x,Y_{x'})$ attainable by PNSCMs and PDSCMs in $\mathfrak M$ coincide. Absent additional structural restrictions, this common set is the full probability simplex over $\{0,1\}^3$.
\end{proposition}

See Appx.~\ref{appx:proofs} for all proofs.
An immediate corollary of Prop.~\ref{thm:inheritance} is that any sharp bound on $\mathrm{PN}$, $\mathrm{PS}$, or $\mathrm{PNS}$ that depends only on the unrestricted joint law of $(X,Y_x,Y_{x'})$ carries over unchanged from PDSCMs to PNSCMs.
However, stochastic mechanisms allow for stochastic constraints that cannot be formalized using PDSCM semantics. We consider several such constraints below.

\section{A generic identification program}
\label{sec:generic_program}

\begin{wrapfigure}[9]{r}{0.38\textwidth}
  \centering
  \vspace{-6mm}
  \begin{tikzpicture}[
    >={Stealth[length=2mm,width=1.6mm]},
    obs/.style={circle,draw,minimum size=9mm,inner sep=0pt},
    lat/.style={circle,draw,dashed,minimum size=9mm,inner sep=0pt},
    pot/.style={circle,draw,fill=black!12,minimum size=9mm,inner sep=0pt},
    arr/.style={->,thick}
  ]

  \node[lat] (U)  at (2.2,3.0) {$U$};

  \node[lat] (A)  at (0,1.5)   {$A$};
  \node[lat] (B)  at (2.2,1.5) {$B$};
  \node[lat] (C)  at (4.4,1.5) {$C$};

  \node[obs] (X)  at (0,0)     {$X$};
  \node[pot] (Yx) at (2.2,0)   {$Y_x$};
  \node[pot] (Yz) at (4.4,0)   {$Y_{x'}$};

  \draw[arr] (U) -- (A);
  \draw[arr] (U) -- (B);
  \draw[arr] (U) -- (C);

  \draw[arr] (A) -- (X);
  \draw[arr] (B) -- (Yx);
  \draw[arr] (C) -- (Yz);

\end{tikzpicture}
  \caption{Modified DAG, replacing $Y$ with potential outcomes (shaded) and introducing latent nodes (dashed) for treatment propensity $A$ and stochastic response variables $B, C$.}
  \label{fig:dag2}
\end{wrapfigure}
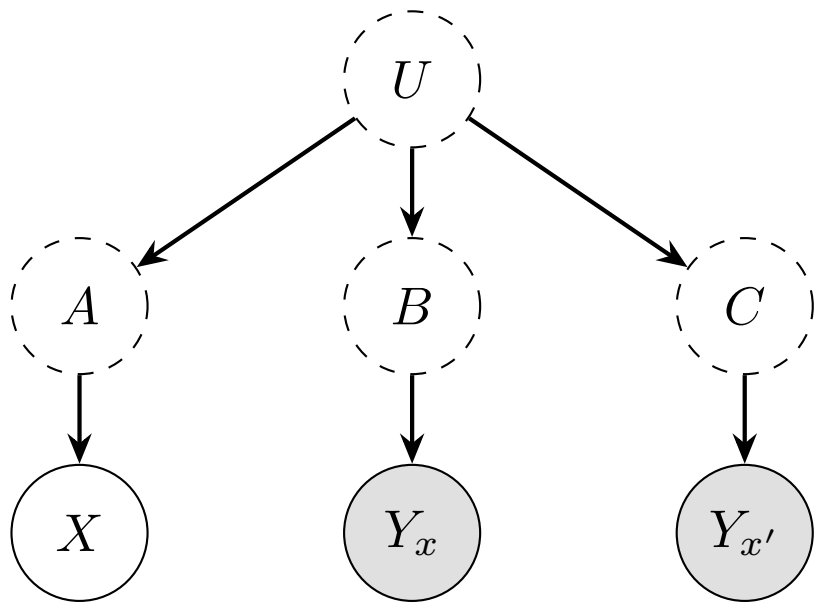

Before introducing specific structural assumptions, we formulate the partial identification problem in a common optimization framework.
We begin by reparameterizing the model with separate variables for each source of latent influence:
\begin{align*}
    \underbrace{A(U):=p(x\mid U)}_{\text{treatment propensity}}, \quad \underbrace{B(U):=p(y\mid x,U), \quad C(U):=p(y\mid x',U)}_{\text{stochastic response variables}}.
\end{align*}
This parameterization works because all observational, interventional, and counterfactual quantities considered below depend on the latent state $u$ only through the triple $(A(u),B(u),C(u))$. 
Redrawing our DAG with potential outcome variables, we see that $A, B$ and $C$ represent the three channels through which latent information flows (see Fig.~\ref{fig:dag2}).
Since these define an equivalence class over latent mechanisms, we henceforth replace the original $U$ with its canonical representation $U:=(A,B,C)\in[0,1]^3$, with corresponding latent law $\mu$.

For a fixed collection of $\mathcal L_1, \mathcal L_2$ distributions $\mathcal D$, compatibility imposes five generalized moment constraints on $\mu$, which we separate into observational and interventional components:
\begin{equation*}
\renewcommand{\arraystretch}{1.6}
\begin{array}{r@{\;}c@{\;}l | r@{\;}c@{\;}l}
\multicolumn{3}{c|}{\textbf{obs}} & \multicolumn{3}{c}{\textbf{int}} \\[2pt]
\hline
p(x) &=& \displaystyle\int a \,d\mu 
    & p(y_x) &=& \displaystyle\int b \,d\mu \\
p(x,y) &=& \displaystyle\int a\,b \,d\mu
    & p(y_{x'}) &=& \displaystyle\int c \,d\mu \\
p(x',y) &=& \displaystyle\int (1-a)\,c \,d\mu
    & \multicolumn{3}{c}{} \\
\end{array}
\end{equation*}
Together with $\int d\mu = 1$, these constraints characterize the latent laws compatible with $\mathcal D$, thereby defining $\mathfrak M(\mathcal D)$.

Next, we define target estimands in terms of our latent variables, observing that each probability of causation is a linear functional of $\mu$ (for an explicit derivation, see Appx. \ref{appx:linear}):
\begin{align*}
    q_{\mathrm{PN}}(a,b,c) := \frac{ab(1-c)}{p(x,y)}, \quad
    q_{\mathrm{PS}}(a,b,c) := \frac{(1-a)b(1-c)}{p(x',y')}, \quad 
    q_{\mathrm{PNS}}(a,b,c) := b(1-c).
\end{align*}
Let $\mathcal P([0,1]^3)$ denote the set of Borel probability measures on the three-dimensional unit cube.
Let $\mathcal C$ denote some additional structural restriction on the canonical latent law, e.g. the exogeneity and monotonicity conditions of \citetalias{tian_pearl_2000}.
Then for $Q\in \{\mathrm{PN},\mathrm{PS}, \mathrm{PNS}\}$, the generic identification task is:
\begin{align}\label{eq:generic_identification_program} 
    \underset{\mu\in\mathcal P([0,1]^3)}{\operatorname{min/max}}
    \quad&
    \int q_Q \,d\mu, \quad 
    \text{\rm subject to}\quad
    \text{\textbf{obs}, \textbf{int}}, \mu \models \mathcal C. \tag{opt}
\end{align}
Observational and interventional distributions therefore determine a generalized moment fiber, while extra assumptions enter through constraint(s) $\mathcal C$.
We say that bounds are \textit{sharp under} $\mathcal C$ if endpoints equal the min/max of the corresponding $Q$ over all models compatible with $\mathcal D$ that satisfy $\mathcal C$. 

This constrained optimization task is closely related to the LP of \citetalias{tian_pearl_2000}---itself adapted from prior work by \citet{balke1994}---who take the latent state to be the deterministic Boolean counterfactual type $U_{\mathrm{TP}}:=(X,Y_x,Y_{x'}) \in\{0,1\}^3$
and optimize directly over the eight cell probabilities $p_{ijk}:=p(X=i,Y_x=j,Y_{x'}=k)$.
Both the constraints and objectives are then linear in $p_{ijk}$, yielding a finite LP. 
In the present setting, $U=(A,B,C)$ instead indexes stochastic mechanisms. 
A single latent type may induce probability on several Boolean counterfactual configurations, with 
$p_{ijk} = \int
    a^i(1-a)^{1-i}~
    b^j(1-b)^{1-j}~
    c^k(1-c)^{1-k}~
    d\mu$.
Absent further restrictions, the inheritance result of
Prop.~\ref{thm:inheritance} implies that both formulations generate the same feasible binary counterfactual region and therefore the same assumption-free bounds. The distinction only becomes clear once restrictions $\mathcal C$ are imposed on latent mechanisms.

Although \eqref{eq:generic_identification_program} is formally an optimization over probability measures, many structural assumptions reduce it exactly to a finite-dimensional problem. 
Call a constraint set $\mathcal C$ \emph{finitely reducible} for the present identification task if there exists some finite $K$ such that every feasible value of $Q$ admits a compatible atomic law $\mu = \sum_{i=1}^K \pi_i~\delta_{(a_i,b_i,c_i)}$, where $\forall i: \pi_i\geq0$ and $\sum_i\pi_i=1$.
Under such a representation, all integrals are replaced with summations over simplex weights $\pi_i$. 
This property holds, in particular, for finite collections of support restrictions and generalized moment constraints.
Below, we focus on assumptions that are natural for nondeterministic causal models and finitely reducible w.r.t. \eqref{eq:generic_identification_program}. 

\section{Mechanism Uncertainty}\label{sect:mechanism}

In this section, we consider a range of structural assumptions on the causal mechanisms for $X \mid U$ and $Y \mid X, U$.
Each assumption constrains the support of our latent variables $A,B,C$, leading to sharp bounds on the probabilities of causation.
All proofs are deferred to Appx. \ref{appx:proofs}.

\subsection{Margin condition}\label{sec:margin}

Our first constraint is a simple positivity condition. 
In a PDSCM, fixing the latent $U=u$ drives conditional probabilities \mbox{$p(x \mid u)$}, \mbox{$p(y\mid x,u)$}, and \mbox{$p(y\mid x',u)$} to some value in $\{0,1\}$. 
A PNSCM, by contrast, naturally accommodates intermediate probabilities.
We therefore consider mechanisms that remain uniformly bounded away from the edges of the unit cube.

\begin{definition}[Margin condition]
For $\varepsilon = (\varepsilon_A, \varepsilon_B, \varepsilon_C) \in[0,1/2)^3$, model $\mathcal M$ satisfies the $\varepsilon$-\textit{margin condition} if, for all $u \in \mathcal U$:
\begin{align*}
    \varepsilon_A \leq p(x \mid u) \leq 1-\varepsilon_A, ~~ \varepsilon_B \leq p(y\mid x,u) \leq 1 - \varepsilon_B, ~~ \varepsilon_C \leq p(y \mid x', u) \leq 1 - \varepsilon_C.
\end{align*}
\end{definition}
Interpreting $p(x \mid u)$ as a latent propensity score, nonzero values of $\varepsilon_A$ amount to a latent analogue of the standard overlap assumption on treatment assignment \citep{rosenbaum1983central, Imbens:2015fk, hernan2025causal}. 
For the response mechanism, nonzero values of $\varepsilon_B, \varepsilon_C$ mean that no latent state guarantees a specific value for either potential outcome variable. 
Geometrically, we can interpret the $\varepsilon$-margin condition as reducing the support for $(A,B,C)$ from the unit cube to the subset $\mathcal R_\varepsilon := [\varepsilon_A, 1 - \varepsilon_A] \times [\varepsilon_B, 1 - \varepsilon_B] \times [\varepsilon_C, 1 - \varepsilon_C]$.
If any margin is nonzero, this becomes an intrinsically stochastic assumption that cannot be satisfied by any PDSCM.

An equivalent way to understand the margin condition is to define the rescaled latent variables $\tilde J := (J - \varepsilon_J)/d_J$, where $d_J := 1 - 2 \varepsilon_J$ for $J \in \{A, B, C\}$.
The margin condition states that $(\tilde A, \tilde B, \tilde C) \in [0,1]^3, \mu$-almost surely.
We use these rescaled latents to define binary variables $(\tilde X, \tilde Y_{\tilde x}, \tilde Y_{\tilde x'})$, such that $p(\tilde x \mid U) = \tilde A(U)$, $p(\tilde y \mid \tilde x, U) = \tilde B(U)$, and $p(\tilde y \mid \tilde x', U) = \tilde C(U)$, as well as the auxiliary distribution $\tilde p_{ijk} := p(\tilde X=i, \tilde Y_{\tilde x} = j, \tilde Y_{\tilde x'} = k)$. 

\begin{proposition}[Testability of $\varepsilon$-margin]
\label{prop:margin_testability}
The $\varepsilon$-margin condition is compatible with $\mathcal D$ iff
there exist simplex weights $\tilde p_{ijk}:i,j,k\in\{0,1\}$ such that:
\begin{gather*}
    \sum_{j,k}\tilde p_{1jk} = \frac{p(x)-\varepsilon_A}{d_A},  \quad
    \sum_{i,k}\tilde p_{i1k} = \frac{p(y_x)-\varepsilon_B}{d_B}, \quad
    \sum_{i,j}\tilde p_{ij1} = \frac{p(y_{x'})-\varepsilon_C}{d_C},\\
    \sum_k\tilde p_{11k} =
    \frac{
        p(x,y)
        -\varepsilon_A p(y_x)
        -\varepsilon_B p(x)
        +\varepsilon_A\varepsilon_B
    }{d_A d_B},
    \label{eq:margin_test_xy}\\
    \sum_j\tilde p_{0j1} =
    \frac{
        p(x',y)
        -\varepsilon_A p(y_{x'})
        -\varepsilon_C p(x')
        +\varepsilon_A\varepsilon_C
    }{d_A d_C}.
\end{gather*}
\end{proposition}

\noindent Prop.~\ref{prop:margin_testability} suggests a data-driven strategy for evaluating feasible values of latent parameters, an observation we exploit in our experiments below.

For $Q \in \{\mathrm{PN}, \mathrm{PS},\mathrm{PNS}\}$ and data $\mathcal D$, define the assumption-free bounds $Q^\pm_\emptyset(\mathcal D)$ as the solutions of \eqref{eq:generic_identification_program} without restriction $\mathcal C$, i.e.\ the extrema of $Q(\mathcal M)$ over $\mathcal M \in \mathfrak M(\mathcal D)$.
With access to $\mathcal L_1$ and $\mathcal L_2$ data, \citetalias{tian_pearl_2000} derive the following extrema using standard Fr\'echet inequalities:
\begin{align*}
    \mathrm{PN}^-_\emptyset &:= \max\left\{0,\, \frac{p(y)-p(y_{x'})}{p(x,y)} \right\}, \quad 
    \mathrm{PN}^+_\emptyset := \min\left\{1,\,\frac{p(y'_{x'})-p(x',y')}{p(x,y)}\right\},\\[1mm]
    \mathrm{PS}^-_\emptyset &:= \max\left\{0,\, \frac{p(y_x)-p(y)}{p(x',y')} \right\}, \quad 
    \mathrm{PS}^+_\emptyset := \min\left\{1,\,\frac{p(y_x)-p(x,y)}{p(x',y')}\right\},\\
    \mathrm{PNS}^-_\emptyset &:= \max\left\{0,\,p(y_x)-p(y_{x'}),\,p(y)-p(y_{x'}),\,p(y_x)-p(y)\right\},\\
    \mathrm{PNS}^+_\emptyset &:= \min\left\{p(y_x),\,p(y'_{x'}),\,p(x,y)+p(x',y'),\,p(y_x)-p(y_{x'})+p(x,y')+p(x',y)\right\}.
\end{align*}

We apply a similar strategy to our modified pmf. To concisely express marginals, write $\tilde p_{i1+} := \sum_k \tilde p_{i1k}$, $\tilde p_{i+0} := \sum_j \tilde p_{ij0}$, and $\tilde p_{i++} := \sum_{j,k} \tilde p_{ijk}$.
Then for $i \in \{0,1\}$, define
\begin{gather*}
    \tilde \ell_i :=
    \max\{
        0,\, \tilde p_{i1+}+\tilde p_{i+0}-\tilde p_{i++} 
    \}, \quad 
    \tilde h_i :=
    \min\{
        \tilde p_{i1+},\tilde p_{i+0}
    \},\\
    f_i(z) :=
    \varepsilon_B\varepsilon_C\tilde p_{i++}
    +\varepsilon_Bd_C\tilde p_{i+0}
    +d_B\varepsilon_C\tilde p_{i1+}
    +d_Bd_Cz.
\end{gather*}
The first two quantities represent Fr\'echet bounds on
$\tilde p_{i10}$, while $f_i(\tilde p_{i10}) = \mathbb E[\mathbb I\{\tilde X=i\}B(1-C)]$ maps the corresponding auxiliary cell back to its contribution on the original response scale.
\begin{theorem}[$\varepsilon$-margin bounds]
\label{thm:margin}
The following bounds are sharp under the $\varepsilon$-margin condition:
\begin{align*}
    \frac{\varepsilon_A f_0(\tilde \ell_0)+(1-\varepsilon_A)f_1(\tilde \ell_1)}{p(x,y)} &\leq \mathrm{PN} \leq \frac{\varepsilon_A f_0(\tilde h_0)+(1-\varepsilon_A)f_1(\tilde h_1)}{p(x,y)}\\
    \dfrac{(1-\varepsilon_A)f_0(\tilde \ell_0)+\varepsilon_A f_1(\tilde \ell_1)}{p(x',y')} &\leq \mathrm{PS} \leq \dfrac{(1-\varepsilon_A)f_0(\tilde h_0)+\varepsilon_A f_1(\tilde h_1)}{p(x',y')}\\
    f_0(\tilde \ell_0) + f_1(\tilde \ell_1) &\leq \mathrm{PNS} \leq f_0(\tilde h_0) + f_1(\tilde h_1).
\end{align*}
\end{theorem}
Thm.~\ref{thm:margin} shows that the margin restriction preserves the basic geometry of the \citetalias{tian_pearl_2000} problem. After rescaling, the remaining cross-world uncertainty is captured by the same pair of Fr\'echet couplings, while $f_i$ and the weights determined by $\varepsilon_A$ map the resulting auxiliary cells back to the original PNSCM.
Accordingly, the margin-constrained intervals are contained in the assumption-free \citetalias{tian_pearl_2000} intervals and reduce exactly to them when $\varepsilon_A = \varepsilon_B = \varepsilon_C = 0$. The containment is strict whenever the $\varepsilon$-margin condition excludes the corresponding \citetalias{tian_pearl_2000} extremizer.

\subsection{Soft exogeneity}
\label{sec:gamma_deviation}

\citetalias{tian_pearl_2000}'s first substantive assumption is what they call \textit{exogeneity}, which occurs when $p(y \mid x) = p(y_x)$ and $p(y \mid x') = p(y_{x'})$---equivalently, when $X \indep Y_x$ and $X \indep Y_{x'}$. This amounts to a ``no confounding'' condition that severs at least one outgoing edge from $U$ in the DAG of Fig.~\ref{fig:dag1}, leading to sharp bounds on PNS:
\begin{align*}
    \mathrm{PNS}_e^- := \max\left\{0, p(y \mid x) - p(y \mid x')\right\}, \quad \mathrm{PNS}_e^+ :=  \min\left\{p(y \mid x), p(y' \mid x')\right\},
\end{align*}
with corresponding bounds on PN and PS:\footnote{The PN bound is originally due to \citet{Robins1989ThePO}; the other bounds were first derived under a slightly stronger condition by \citet{pearl99}.}
\begin{align*}
    \mathrm{PN}_e^\pm := \frac{\mathrm{PNS}_e^\pm}{p(y \mid x)}, \quad
    \mathrm{PS}_e^\pm := \frac{\mathrm{PNS}_e^\pm}{p(y' \mid x')}.
\end{align*}
These are ordinary Fr\'echet bounds on the unresolved cross-world coupling between $Y_x$ and $Y_{x'}$. 

We introduce a quantitative restriction that permits bounded deviations from exogeneity. 
\begin{definition}[$\delta$-exogeneity]
    For $\delta \in [0, 1]$, $\mathcal M$ satisfies $\delta$-exogeneity if, for all $u \in \mathcal U$: 
    \begin{align*}
        |p(x \mid u) - p(x)| \leq \delta.
    \end{align*}
\end{definition}
\noindent The condition may be interpreted as a form of \textit{soft exogeneity}, since it limits variation in treatment propensity.
At $\delta=0$, we sever the $U \rightarrow X$ edge altogether, effectively randomizing treatment. 
Larger values of $\delta$ gradually allow for greater confounding.
For sufficiently large values of $\delta$, the restriction becomes vacuous and the assumption-free \citetalias{tian_pearl_2000} region is recovered.
Such parameterized relaxations of causal assumptions are common in the literature on instrumental variables \citep{Conley2012Plausibly, silva_evans:16, watson2024bounding}.

Note that $\delta$-exogeneity specifically targets the treatment assignment mechanism. 
We could alternatively limit confounding by targeting the $U \rightarrow Y$ relationship. 
However, it is often easier to (approximately) randomize treatments than it is to (approximately) randomize response mechanisms, as the former are at least partially under researcher control. 
This is the guiding principle behind intention-to-treat (ITT) analysis \citep{Gupta2011IntentionToTreat} and related methods for analyzing data from studies with imperfect compliance \citep{balke97}.

Define $\underline a := \max\{0, p(x) - \delta\}$ and $\overline a := \min\{1, p(x) + \delta\}$.
Then $\delta$-exogeneity reduces the support of our latent variables $(A, B, C)$ from the unit cube to $\mathcal R_\delta := [\underline a, \overline a] \times [0,1]^2$. This leads to the following test. 
\begin{proposition}[Testability of $\delta$-exogeneity]
\label{prop:delta_testability}
The $\delta$-exogeneity condition is compatible with $\mathcal D$ iff
\begin{align*}
    \max\left\{
        \underline a\,p(y_x),\,
        p(x)-\overline ap(y'_x)
    \right\} &\leq p(x,y) \leq \min\left\{
        \overline a\,p(y_x),\,
        p(x)-\underline a p(y'_x)
    \right\}, \\
    \max\left\{
        \underline a\,p(y_{x'}),\,
        p(x)-\overline a p(y'_{x'})
    \right\} &\leq p(y_{x'})-p(x',y) \leq \min\left\{
        \overline a\,p(y_{x'}),\,
        p(x)-\underline a p(y'_{x'})
    \right\}.
\end{align*}
\end{proposition}

\noindent As with Prop.~\ref{prop:margin_testability},  Prop.~\ref{prop:delta_testability} suggests a method for computing feasible values for latent parameters.

For $\delta>0$, let $d := \overline a - \underline a, A_\delta := (A - \underline a)/d$, and define an auxiliary treatment variable $X_\delta$ such that $p(x_\delta \mid U) = A_\delta(U)$. 
Let $p^\delta_{ijk} = p(X_\delta=i, Y_x = j, Y_{x'}=k)$ denote an auxiliary pmf with relevant marginals identified via:
\begin{align*}
    p^\delta_{1++} = \frac{p(x) - \underline a}{d}, \quad 
    p^\delta_{11+} = \frac{p(x, y) - \underline a p(y_x)}{d}, \quad
    p^\delta_{1+1} = \frac{p(y_{x'}) - p(x', y) - \underline a p(y_{x'})}{d},
\end{align*}
with corresponding $i=0$ marginals following by subtraction from $p(y_x), p(y_{x'})$, and $1$. 
Then sharp bounds on $p^\delta_{i10}$ are given by $\ell^\delta_i := \max\{0, p^\delta_{i1+} + p^\delta_{i+0} - p^\delta_{i++}\}$ and $h^\delta_i := \min\{p^\delta_{i1+}, p^\delta_{i+0}\}$,
leading to the following result.
\begin{theorem}[$\delta$-exogeneity bounds]
\label{thm:delta_exogeneity}
The following bounds are sharp under $\delta$-exogeneity:
\begin{align*}
    \frac{
        \underline a\,\ell^\delta_0
        +\overline a\,\ell^\delta_1
    }{p(x,y)}
    &\leq
    \mathrm{PN}
    \leq
    \frac{
        \underline a\,h^\delta_0
        +\overline a\,h^\delta_1
    }{p(x,y)},
    \\[1mm]
    \frac{
        (1-\underline a)\ell^\delta_0
        +(1-\overline a)\ell^\delta_1
    }{p(x',y')}
    &\leq
    \mathrm{PS}
    \leq
    \frac{
        (1-\underline a)h^\delta_0
        +(1-\overline a)h^\delta_1
    }{p(x',y')},
    \\[1mm]
    \ell^\delta_0+\ell^\delta_1
    &\leq
    \mathrm{PNS}
    \leq
    h^\delta_0+h^\delta_1.
\end{align*}
\end{theorem}

Thm.~\ref{thm:delta_exogeneity} requires $\delta>0$, since $d=0$ at $\delta=0$. At $\delta=0$, compatibility requires exact exogeneity and the sharp bounds are $Q_e^\pm$. These are also the limit as $\delta \downarrow 0$: for exogenous data, the bounds of Thm.~\ref{thm:delta_exogeneity} coincide with $Q_e^\pm$ for every $\delta>0$.
For $\delta\geq\max\{p(x),p(x')\}$, the restriction is vacuous and the bounds reduce to the assumption-free \citetalias{tian_pearl_2000} bounds $Q_\emptyset^\pm$.
For intermediate $\delta$, the sharp region is obtained by applying the same Fr\'echet construction to the rescaled treatment distribution $p^\delta$.

\subsection{Stochastic monotonicity}\label{sec:smon}

Classical monotonicity rules out the joint counterfactual event $y'_x \wedge y_{x'}$. Under PNSCM semantics, this is equivalent to $p(y' \mid x,u)\,p(y\mid x',u)=0$, for all $u \in \mathcal U$.
Thus, at every latent state, classical monotonicity requires either $p(y\mid x,u)=1$ or $p(y\mid x',u)=0$. Such extreme-valued probabilities are a demanding restriction on genuinely stochastic mechanisms. We therefore consider the following weaker condition.

\begin{definition}[Stochastic monotonicity]
    $Y$ is stochastically monotone (s-monotone) in $X$ if, for all $u \in \mathcal U: p(y\mid x,u)\geq p(y\mid x',u)$.
\end{definition}
Stochastic monotonicity says that treatment cannot reduce the probability of $y$ at any latent state. Unlike classical monotonicity, it permits $p(y\mid x,u)$ and $p(y\mid x',u)$ to simultaneously lie strictly between zero and one.
A related probabilistic relaxation of monotonicity was proposed by \citet{ramsahai2012} for instrumental variable models, where latent treatment probabilities are ordered across instrument levels. 
Geometrically, we can think of s-monotonicity as reducing the feasible region for $(B,C)$ from the unit square to an isosceles right triangle, giving the feasible polytope $\mathcal R_s := [0,1] \times \left\{(b, c): 0 \leq c \leq b \leq 1\right\}$.

S-monotonicity has a directly testable implication.
\begin{proposition}[Testability of s-monotonicity]
\label{prop:smono_testability}
The s-monotonicity condition is compatible
with $\mathcal D$ iff $p(y_x)\geq p(y)\geq p(y_{x'})$.
\end{proposition}
Note that this same condition, applied under PDSCM semantics, implies classical monotonicity. 
In that case, \citetalias{tian_pearl_2000} show that probabilities of causation are identifiable when observational and interventional data are available: $\mathrm{PN}_{m} := (p(y) - p(y_{x'}))/p(x, y)$, $\mathrm{PS}_{m} := (p(y_x) - p(y))/p(x', y')$, and $\mathrm{PNS}_{m} := p(y_x) - p(y_{x'})$.
By contrast, s-monotonicity generally yields partial identification intervals.

\begin{theorem}[S-monotonicity bounds]
\label{thm:smon}
The following bounds are sharp under s-monotonicity: 
\begin{gather*}
    \mathrm{PN}_{m} \leq \mathrm{PN} \leq \mathrm{PN}_{m} + \psi_{\mathrm{PN}}, \quad
    \mathrm{PS}_{m} \leq \mathrm{PS} \leq \mathrm{PS}_{m} + \psi_{\mathrm{PS}},\\
    \mathrm{PNS}_{m} \leq \mathrm{PNS} \leq \mathrm{PNS}_{m} + \psi_{\mathrm{PNS}},
\end{gather*}
with stochastic slack terms:
\begin{gather*}
    \psi_{\mathrm{PN}} := (1 - \mathrm{PN}_{m}) \frac{p(x, y')}{p(x, y') + p(y_{x'}) - p(x', y)}, \quad
    \psi_{\mathrm{PS}} := (1 - \mathrm{PS}_{m}) \frac{p(x', y)}{p(x') - p(y_x) + p(y)},\\
    \psi_{\mathrm{PNS}} := p(x, y) \psi_{\mathrm{PN}} + p(x', y') \psi_{\mathrm{PS}}.
\end{gather*}
\end{theorem}

The lower endpoint of each interval is exactly the point-identification formula obtained by \citetalias{tian_pearl_2000} under classical monotonicity.
The upper endpoints add a nonnegative \emph{stochastic slack} term, which captures the additional probability of causation that can arise when treatment increases the probability of $y$ at every latent state without deterministically fixing potential outcomes.

\section{Case Study: Personalized Treatment Decisions}\label{sect:application}
\defcitealias{dawid_senn_2025}{DS}
\defcitealias{sarvet_stensrud_2025}{SS}
Probabilities of causation play a central role in a recent debate on personalized decision-making in medicine. 
Let $x$ denote treatment and $y$ recovery. 
\citet{mueller2023} argue that the decision whether to treat a patient should take into account the probability of benefit, $P(B) := \mathrm{PNS}$, and the probability of harm, $P(H) := p(y'_x, y_{x'})$, rather than just the average treatment effect, $\mathrm{ATE} := p(y_x) - p(y_{x'}) = P(B) - P(H)$.
\citet[henceforth \citetalias{dawid_senn_2025}]{dawid_senn_2025} instead defend the standard rule, treat iff $\mathrm{ATE} > 0$, and \citet[henceforth \citetalias{sarvet_stensrud_2025}]{sarvet_stensrud_2025} raise related objections. 
Following \citetalias{dawid_senn_2025} and \citet{beckers2026harm}, the former proposal can be read as the rule
\begin{align}\label{eq:rule}
    \text{treat} \iff P(B) > (1+\lambda)\,P(H) \iff P(H) < \mathrm{ATE}/\lambda,
\end{align}
for a harm aversion coefficient $\lambda \geq 0$. This amounts to maximizing the harm-penalized utility of \citet{richens2022} with default action $x'$, and reduces to the standard rule for $\lambda = 0$. 

\citetalias{dawid_senn_2025} and \citetalias{sarvet_stensrud_2025} offer a twofold critique of rule \eqref{eq:rule}. 
First, they criticize the underlying notion of harm, which takes withholding treatment as the default action, on philosophical and ethical grounds. 
Second, they criticize its limited applicability: unless one makes unrealistic assumptions, $P(B)$ and $P(H)$ remain undetermined, so the rule is mostly uninformative. 
We set aside the first criticism altogether, and refer the reader to \citet{beckers2024causal, beckers2023harm, beckers2026harm} for an in-depth discussion.
Our focus lies with the second criticism.

Here it is essential to distinguish between identifying $P(H)$ and deciding whether rule \eqref{eq:rule} applies. 
The latter only requires an upper bound $H^+ < \mathrm{ATE}/\lambda$ to justify treatment, or a lower bound $H^- \geq \mathrm{ATE}/\lambda$ to rule it out. 
Let $\lambda^* := \mathrm{ATE}/H^+$ denote the largest harm aversion for which treatment is thereby justified.
Both \citetalias{dawid_senn_2025} and \citetalias{sarvet_stensrud_2025} focus on the former, arguing that partial identification introduces ambiguities that must be resolved by other means, e.g. taking midpoints or minimizing regret.
Classical monotonicity gives $H^+ = 0$, and thus collapses \eqref{eq:rule} into the standard rule, but only by requiring extreme probabilities at every latent state (Sect.~\ref{sec:smon}).
Otherwise, writing $b_z := p(y_x \mid z)$ and $c_z := p(y_{x'} \mid z)$ for $z \in \{x, x'\}$---which are identifiable from $\mathcal D$ via consistency and cross-world equations (see Eq.~\eqref{eq:crossworld}, Appx.~\ref{appx:proofs})---$P(H)$ is point identified only if, for each $z$, some $b_z, c_z$ lies in $\{0,1\}$, ``a most unusual state of affairs'' \citepalias[Remarks~1--2]{dawid_senn_2025}.
Yet for the very data that \citetalias{dawid_senn_2025} use in their Example~4 to illustrate that $P(H)$ remains unidentified even when observational data are added, the assumption-free bounds already establish that rule \eqref{eq:rule} justifies treatment for $\lambda = 2$, the value they adopt (writing \eqref{eq:rule} as $P(B) > 3P(H)$), since $\lambda^* = 2.33$ (see Appx.~\ref{appx:application}).

S-monotonicity makes it easier to decide whether rule \eqref{eq:rule} applies, by lowering the upper bound on $P(H)$ without positing any extreme response probabilities.

\begin{table}[t]
\centering
\small
\caption{Sharp bounds on the probabilities of benefit and harm in the example, the resulting decision under rule \eqref{eq:rule} with $\lambda = 2$, and the largest harm aversion $\lambda^*$ for which treatment is justified.}
\begin{tabular}{lcccc}
\hline
Assumptions & $P(B)$ & $P(H)$ & $\lambda = 2$ & $\lambda^*$ \\
\hline
Experimental data only        & $[0.28, 0.49]$  & $[0, 0.21]$  & undecided & $1.33$ \\
$\mathcal D$, no restriction  & $[0.28, 0.44]$  & $[0, 0.16]$  & undecided & $1.75$ \\
$\mathcal D$, s-monotonicity  & $[0.28, \bm{0.395}]$ & $[0, \bm{0.115}]$ & treat     & $2.44$ \\
$\mathcal D$, monotonicity    & $0.28$          & $0$          & treat     & $\infty$ \\
\hline
\end{tabular}
\label{tab:decision}
\end{table}

\begin{corollary}[Harm bounds]
\label{cor:harm}
Let $\tau := \sum_{z} p(z) \min\{c_z, 1 - b_z\}$ and suppose that $p(y_x) \geq p(y) \geq p(y_{x'})$. Then the following bounds are sharp, absent further restrictions and under s-monotonicity, respectively: $0 \leq P(H) \leq \tau$ and $0 \leq P(H) \leq \psi_{\mathrm{PNS}}$. Moreover, $\tau/2 \leq \psi_{\mathrm{PNS}} \leq \min\{\tau, (1 - \mathrm{ATE})/4\}$.
\end{corollary}

The condition of Cor.~\ref{cor:harm} is the test of Prop.~\ref{prop:smono_testability}, and is equivalent to $b_z \geq c_z$ for both $z$ (see Appx.~\ref{appx:smon}), i.e.\ to nonnegative ATE among the treated and the untreated.
By \citetalias{sarvet_stensrud_2025} \citeyearpar[Prop.~1]{sarvet_stensrud_2024}, it thus holds exactly when neither a counterfactual analysis (via sharp bounds on $P(H)$) nor an interventionist analysis (via these conditional effects) detects harm. 
Accordingly, the lower bound on $P(H)$ is zero, so rule \eqref{eq:rule} either justifies treatment or remains undecided.
The key gain lies in the upper bound, which s-monotonicity lowers from $\tau$ to $\psi_{\mathrm{PNS}}$, i.e.\ by up to a factor of two, thereby raising $\lambda^*$. Moreover, s-monotonicity justifies treatment when $\mathrm{ATE} > \lambda/(\lambda+4)$, a criterion that requires only experimental data. Observational data matter only below this threshold.

\paragraph{Example.}
Consider the trial used throughout this debate (\citealp{mueller2023}; \citetalias{dawid_senn_2025}; \citetalias{sarvet_stensrud_2025}, \citeyear{sarvet_stensrud_2024}), with $p(y_x) = 0.49$ and $p(y_{x'}) = 0.21$, so $\mathrm{ATE} = 0.28$. 
Suppose that, in contrast to Example~4 of \citetalias{dawid_senn_2025}, an observational study of the same population yields $p(x, y) = 0.28$, $p(x, y') = 0.42$, $p(x', y) = 0.14$, and $p(x', y') = 0.16$. 
Then $(b_x, c_x, b_{x'}, c_{x'}) = (0.40, 0.10, 0.70, 0.47)$, so $P(H)$ is not point identified, while s-monotonicity is compatible with $\mathcal D$ since $p(y_x) = 0.49 \geq p(y) = 0.42 \geq p(y_{x'}) = 0.21$.
Thm.~\ref{thm:smon} gives $\psi_{\mathrm{PNS}} = 0.28 \cdot 0.214 + 0.16 \cdot 0.342 = 0.115$.
For $\lambda = 2$, treatment requires $P(H) < 0.14$, which experimental data alone cannot establish since $\mathrm{ATE} < 1/3$.
As Table~\ref{tab:decision} shows, s-monotonicity settles a decision that the assumption-free bounds leave open, without positing any extreme response probabilities. 

The value $\lambda = 2$ was adopted with deterministic response types in mind. In a PNSCM, $P(H)$ also counts harm events arising from intrinsic outcome variation, and is positive at almost every latent state, even when treatment raises the probability of recovery everywhere (Appx.~\ref{appx:application}). A $\lambda$ chosen for deterministic models is thus more demanding than intended, so evaluating rule~\eqref{eq:rule} at $\lambda = 2$ is an a fortiori conservative test: recalibrating $\lambda$ downward only makes $P(H) < \mathrm{ATE}/\lambda$ easier to satisfy, and leaves $\lambda^*$ unaffected. A decision-maker could anchor $\lambda$ on the level of $P(H)$ produced by intrinsic variation alone, e.g.\ by an ineffective treatment with $B(u) = C(u)$, for which $P(H \mid u) = C(u)(1 - C(u))$.

\section{Experiments}

In this section, we apply our parameterized margin and soft exogeneity bounds in semi-synthetic experiments.
This ensures access to ground truth while maintaining realistic joint distributions over latents and observables.
Further details can be found in Appx.~\ref{appx:experiments}, along with more comprehensive simulation benchmarks. Code for reproducing all results is available in the supplement.

We begin by fitting a simple causal model to features from the \texttt{adult} dataset \citep{becker1996adult}. We are interested in the effect of education status $(X)$ on income $(Y)$, controlling for age (presumed to be a common cause of both). 
All variables are binarized---age at $45$, education at bachelor's level, and income at $\$50$k. 
We fit a PNSCM using empirical frequencies with Laplace smoothing $(n=48842)$ to derive $\mathcal L_1, \mathcal L_2$ data. 
Age is subsequently removed from the analysis, introducing latent confounding between $X$ and $Y$.
We compute bounds under a margin condition with uniform $\varepsilon = \varepsilon_A = \varepsilon_B = \varepsilon_C$. 

\begin{figure}[t]
    \centering
    \includegraphics[width=0.88\linewidth]{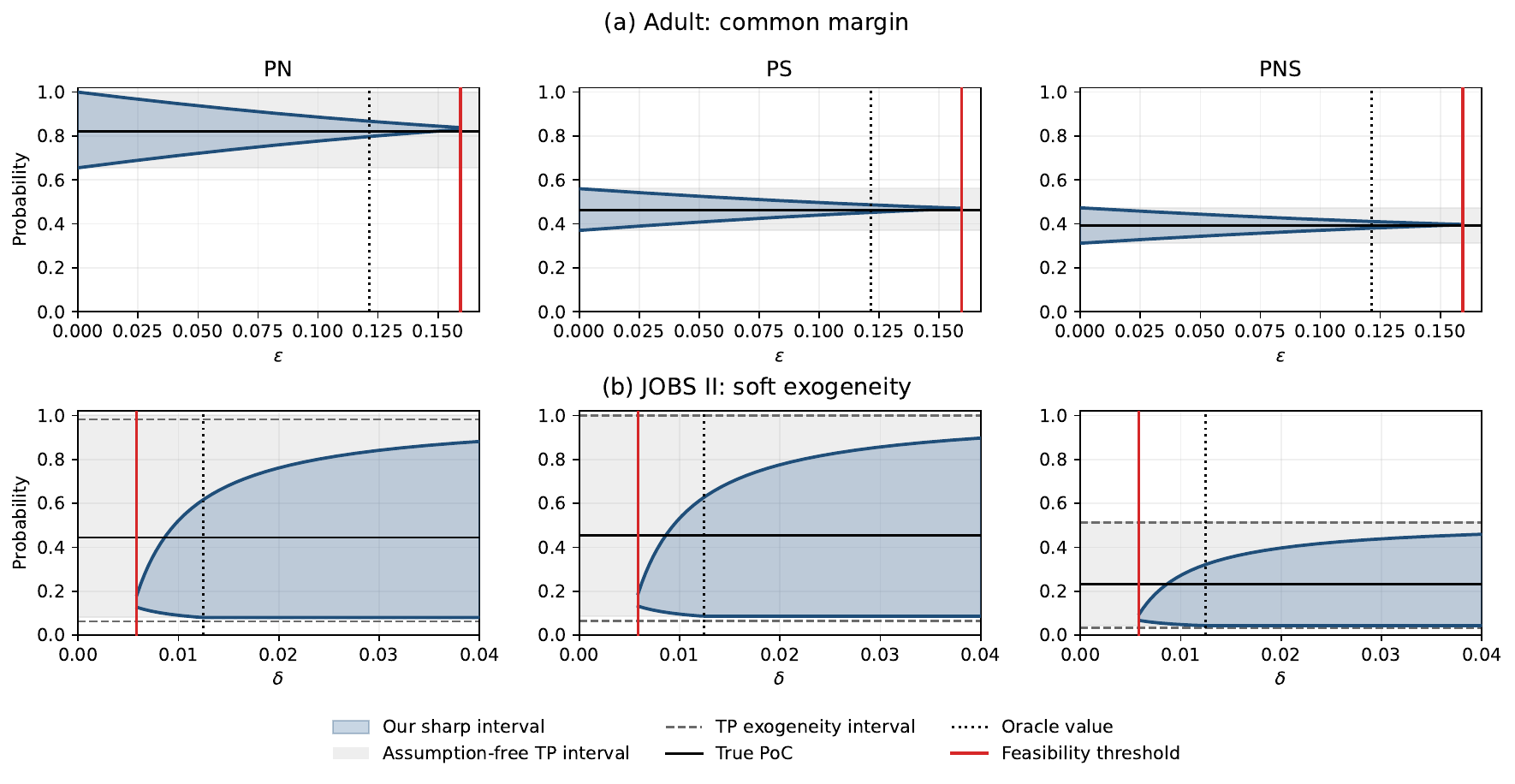}
    \caption{Semi-synthetic experiment. (a) Results with a variable (uniform) margin parameter $\varepsilon$ on the \texttt{adult} dataset. (b) Results with variable soft exogeneity parameter $\delta$ on the JOBS II dataset.}
    \label{fig:semi_synthetic}
\end{figure}

Results are reported in Fig.~\ref{fig:semi_synthetic}(a).
At the oracle value \(\varepsilon^* \approx 0.122\), the resulting intervals are much narrower than the assumption-free \citetalias{tian_pearl_2000} bounds: interval width falls from \(0.344\) to \(0.069\) for PN, from \(0.190\) to \(0.034\) for PS, and from \(0.161\) to \(0.030\) for PNS, corresponding to reductions of roughly $80\%$ each. In this application, the margin restriction yields substantial identification gains while remaining feasible over a broad range of \(\varepsilon\)-values up to approximately \(0.159\).
Even at more conservative margins of $0.1$ or $0.05$, gains remain significant. 
Since the true confounding is presumably richer than binary age, and refining the latent variable can only move mechanism probabilities closer to the boundary, $\varepsilon^*$ should be read as an optimistic value; the bounds at smaller margins indicate what remains under more realistic latent structure.

Next, we analyze data from JOBS II \citep{vinokur1995impact}, 
a randomized trial that evaluated a job-search program for unemployed workers in Michigan. 
The treatment is an offer to participate in the program $(X)$, with measured outcomes including employment status and depressive symptoms based on the Hopkins Symptom Checklist $(Y)$. 
The study has a natural ITT design, with workshop attendance of $54\%$ among offer recipients $(n_{\mathrm{treat}}=1249)$ and $0\%$ for a control group that did not receive the offer $(n_{\mathrm{control}}=552)$.
We binarize the depression variable by thresholding at the sample median, using baseline depression as a latent confounder.

The resulting fitted PNSCM yields a strong illustration of soft exogeneity (see Fig.~\ref{fig:semi_synthetic}(b)). The assumption-free \citetalias{tian_pearl_2000} bounds are almost entirely uninformative for PN and PS, covering nearly the entire unit interval, and remain quite wide for PNS. By contrast, our soft-exogeneity bounds tighten substantially at the oracle value \(\delta^* \approx 0.0125\), reducing interval width by about $40\%$ for all probabilities of causation. The figure also reports the feasibility threshold \(\delta_{\min} \approx 0.0058\), below which the assumption is incompatible with the observed \(\mathcal L_1\) and \(\mathcal L_2\) quantities. For reference we also show the classical TP exogeneity bounds, but these rely on the knife-edge restriction \(\delta=0\); the fitted semi-synthetic model has nonzero oracle \(\delta\), so strict exogeneity is falsified here. Overall, the JOBS II analysis suggests that modest departures from random treatment assignment can still leave enough near-exogeneity to produce meaningful gains in partial identification.

\section{Discussion}\label{sect:discussion}

We have shown that there are tangible benefits to treating causal mechanisms as irreducibly stochastic, even in semi-Markovian models. 
Beyond philosophical motivations, PNSCMs have an expressive advantage over deterministic alternatives, allowing users to formulate reasonable assumptions that are impossible to express in Pearlian semantics. 
These can lead to strictly tighter bounds under a margin condition or soft exogeneity.
Our s-monotonicity results have direct consequences for the ongoing debate over personalized decision-making in medicine, justifying treatment decisions that assumption-free bounds leave open without the extreme probabilities required by determinism.
For each assumption, we characterize compatibility with given $\mathcal L_1$ and $\mathcal L_2$ quantities.

One notable limitation of our margin condition and soft exogeneity assumption is that misspecification of latent parameters can result in invalid bounds that fail to contain the target counterfactual.
This is because our testability conditions (Props. \ref{prop:margin_testability}, \ref{prop:delta_testability}) only evaluate whether \textit{some} admissible model accommodates given values for latent parameters, not whether the true model in fact does. 
One solution is to rely on background knowledge to make informed (ideally, conservative) guesses about plausible values for $\varepsilon, \delta$. A more systematic approach is to report a range of bounds as in Fig.~\ref{fig:semi_synthetic}, allowing users to see how partial identification intervals vary under different assumptions and turning the problem into a sort of sensitivity analysis.


Notable directions for future work include moving from Boolean to multi-valued variables and reducing $\mathcal D$ to just observational quantities, as interventional data is not always available in practice. 
Another is to go beyond the simple two-node graph. 
\citet{lally2026robust} take a first step by formalising counterfactual policy optimisation for MDPs modelled as PNSCMs, bounding the influence of an unobserved global confounder on the transitions by an odds-ratio sensitivity parameter. 
Their setting is more general than ours in that the causal graph unrolls over many states, actions, and time steps, but much more restrictive in that it involves a single global confounder (binary in their application) and none of our structural assumptions. 
Whether our results can be exploited in the sequential setting, e.g.\ to tighten the counterfactual transition bounds that drive robust policy selection, is an interesting question for follow-up work.
A further proposal is to consider constraints on not just the \textit{support} of $\mu$, as we have here, but on other summary statistics like variance or entropy, as in \citep{jiang2023}. 
Analytic bounds may then be unavailable, but numerical optimization should remain feasible.
Such approaches could naturally handle combinations of constraints, such as $\delta$-exogeneity, s-monotonicity, and a Shannon entropy bound.

%% file: appx_CLeaR.tex
\section{Proofs}\label{appx:proofs}
This appendix contains proofs for all propositions and theorems presented in the main text, along with supplemental theoretical results.

We begin by pointing out one important pair of cross-world quantities that are identified by $\mathcal L_1, \mathcal L_2$ data:
\begin{align}\label{eq:crossworld}
    p(y_{x} \mid x') = \frac{p(y_x)-p(x,y)}{p(x')}, \quad 
    p(y_{x'} \mid x) = \frac{p(y_{x'})-p(x',y)}{p(x)}. \tag{CW}
\end{align}
These identities will be exploited repeatedly below.

\subsection{Inheritance}\label{appx:inheritance}

\begin{proof}[Proof of Prop.~\ref{thm:inheritance}]
Let
\begin{align*}
    p_{ijk}
    :=
    p(X=i,Y_x=j,Y_{x'}=k),
    \qquad i,j,k\in\{0,1\},
\end{align*}
denote the joint counterfactual law induced by an arbitrary PNSCM.
Clearly,
\begin{align*}
    p_{ijk}\geq 0,
    \qquad
    \sum_{i,j,k}p_{ijk}=1,
\end{align*}
so every PNSCM induces a point in the probability simplex $\Delta_7$.

We now construct a PDSCM inducing the same point. Introduce a discrete latent variable
\begin{align*}
    U^\star\in\{u_{ijk}:i,j,k\in\{0,1\}\}
\end{align*}
with
\begin{align*}
    p(U^\star=u_{ijk})=p_{ijk}.
\end{align*}
Define deterministic structural functions by
\begin{align*}
    X(u_{ijk}):=i, \quad
    Y(x,u_{ijk}):=j, \quad
    Y(x',u_{ijk}):=k.
\end{align*}
Hence, conditional on $U^\star=u_{ijk}$,
\begin{align*}
    (X,Y_x,Y_{x'})=(i,j,k)
\end{align*}
with probability one. Marginalizing over $U^\star$ therefore gives
\begin{align*}
    p(X=i,Y_x=j,Y_{x'}=k)=p_{ijk}
\end{align*}
for every $(i,j,k)\in\{0,1\}^3$. Thus every joint counterfactual law induced by a PNSCM is also induced by a PDSCM.

Conversely, every PDSCM is a special case of a PNSCM whose conditional mechanisms assign probabilities only in $\{0,1\}$. Hence every joint law induced by a PDSCM is also induced by a PNSCM.

Finally, the construction above applies to an arbitrary $p_{ijk} \in \Delta_7$. 
Therefore both model classes induce exactly the full probability simplex over $(X,Y_x,Y_{x'})$.
\end{proof}


\subsection{Linear representation}\label{appx:linear}

We show that each probability of causation is a linear functional of our latent variables.
Recall the canonical mechanism coordinates
\begin{align*}
    A(U):=p(x\mid U), ~~
    B(U):=p(y\mid x,U), ~~
    C(U):=p(y\mid x',U).
\end{align*}
We first derive the conditional joint law of $(X, Y_x, Y_{x'})$ induced by actualized refinement and conditional product coupling.
Fix a latent state $u$ and write $\bar z$ for the treatment value other than $z \in \{x, x'\}$.
An actual world $(u, z, y_0)$ has probability $p(z \mid u)\, p(Y = y_0 \mid z, u)$.
Under actualized refinement, the counterfactual mechanism for $Y$ returns $y_0$ with certainty at the actual parent configuration $(z, u)$, and coincides with $p(Y \mid \bar z, u)$ at the alternative configuration $(\bar z, u)$.
Hence $Y_z = y_0 = Y$ (consistency), while $Y_{\bar z} \sim p(Y \mid \bar z, u)$, and product coupling makes the two independent given the actual world.
Marginalizing over the factual outcome $y_0 \sim p(Y \mid z, u)$ therefore gives, for all values $j, k$ of $Y$,
\begin{align*}
    p(X = z, Y_x = j, Y_{x'} = k \mid u) = p(z \mid u)\, p(Y = j \mid x, u)\, p(Y = k \mid x', u),
\end{align*}
i.e.\ $X$, $Y_x$ and $Y_{x'}$ are mutually independent given $U = u$, with $p(y_x \mid u) = B(u)$ and $p(y_{x'} \mid u) = C(u)$.
With no latent variables, this yields the point-identification formulas of Sect.~\ref{sect:background}; e.g.\ $\mathrm{PN} = p(x, y, y'_{x'})/p(x, y) = p(y' \mid x')$.
Applying this factorization to the probabilities of causation, first take the PN numerator:
\begin{align*}
    p(x, y_x, y'_{x'} \mid u) &= 
    p(x \mid u) ~p(y_x \mid u) ~p(y'_{x'} \mid u)\\
    &= A(u) ~B(u) ~(1 - C(u)). 
\end{align*}
A similar strategy applies to the PS numerator:
\begin{align*}
    p(x', y_x, y'_{x'} \mid u) &= 
    p(x' \mid u) ~p(y_x \mid u) ~p(y'_{x'} \mid u)\\
    &= (1 - A(u)) ~B(u) ~(1 - C(u)). 
\end{align*}
PNS is simply:
\begin{align*}
    p(y_x, y'_{x'} \mid u) &= p(y_x \mid u) ~p(y'_{x'} \mid u)\\
    &= B(u) ~(1 - C(u)).
\end{align*}
Marginalizing over $U$ therefore gives
\begin{align*}
    p(x,y_x,y'_{x'}) &= \int ab(1-c) ~d\mu\\
    p(x', y_x, y'_{x'}) &= \int (1 - a) b (1 - c) ~d\mu\\
    p(y_x, y'_{x'}) &= \int b(1 - c) ~d\mu.
\end{align*}
Dividing the PN numerator by $p(x, y)$ and the PS numerator by $p(x', y')$ completes the derivation.


\subsection{Margin condition}\label{appx:margin}

\begin{proof}[Proof of Prop.~\ref{prop:margin_testability}]
Suppose that $\mathcal M$ satisfies the margin condition. Writing
\begin{align*}
    A=\varepsilon_A+d_A\tilde A, ~~
    B=\varepsilon_B+d_B\tilde B, ~~
    C=\varepsilon_C+d_C\tilde C,
\end{align*}
we have $(\tilde A,\tilde B,\tilde C)\in[0,1]^3, \mu$-almost surely.
Under the PNSCM product semantics, these coordinates induce the joint law
\begin{align*}
    \tilde p_{ijk} &:=
    p(\tilde X=i,\tilde Y_{\tilde x}=j,
      \tilde Y_{\tilde x'}=k)\\
    &=
    \mathbb E\left[
        \tilde A^i(1-\tilde A)^{1-i}
        \tilde B^j(1-\tilde B)^{1-j}
        \tilde C^k(1-\tilde C)^{1-k}
    \right].
\end{align*}
Hence $\tilde p_{ijk}\geq0$ and $\sum_{ijk}\tilde p_{ijk}=1$.

The first three constraints follow immediately from
\begin{align*}
    \mathbb E[\tilde A] = \frac{p(x)-\varepsilon_A}{d_A}, ~~
    \mathbb E[\tilde B] = \frac{p(y_x)-\varepsilon_B}{d_B}, ~~
    \mathbb E[\tilde C] = \frac{p(y_{x'})-\varepsilon_C}{d_C}.
\end{align*}
Moreover,
\begin{align*}
    p(x,y) &=\mathbb E[AB]\\
    &=
    \varepsilon_A p(y_x) +\varepsilon_B p(x) -\varepsilon_A\varepsilon_B +d_A d_B\mathbb E[\tilde A\tilde B],
\end{align*}
so that
\begin{align*}
    \mathbb E[\tilde A\tilde B] = \sum_k\tilde p_{11k}
    = \frac{p(x,y) -\varepsilon_A p(y_x) -\varepsilon_Bp(x) +\varepsilon_A\varepsilon_B}{d_A d_B}.
\end{align*}
Similarly,
\begin{align*}
    p(x',y) &=\mathbb E[(1-A)C]\\
    &= \varepsilon_Ap(y_{x'}) +\varepsilon_Cp(x') -\varepsilon_A\varepsilon_C +d_A d_C\mathbb E[(1-\tilde A)\tilde C],
\end{align*}
so that 
\begin{align*}
    \mathbb E[(1 - \tilde A) \tilde C] = \sum_j \tilde p_{0j1} = \frac{p(x', y) - \varepsilon_A p(y_{x'}) - \varepsilon_C p(x') + \varepsilon_A \varepsilon_C}{d_A d_C}.
\end{align*}

Conversely, suppose that a pmf $\{\tilde p_{ijk}\}$ satisfying the stated constraints exists.
Let $U$ have states $u_{ijk}$ with masses $\tilde p_{ijk}$ and set
\begin{align*}
    \tilde A(u_{ijk})&=i,&
    \tilde B(u_{ijk})&=j,&
    \tilde C(u_{ijk})&=k.
\end{align*}
Define
\begin{align*}
    A&=\varepsilon_A+d_A\tilde A,&
    B&=\varepsilon_B+d_B\tilde B,&
    C&=\varepsilon_C+d_C\tilde C.
\end{align*}
Then
\begin{align*}
    A \in[\varepsilon_A,1-\varepsilon_A], ~~
    B \in[\varepsilon_B,1-\varepsilon_B], ~~
    C \in[\varepsilon_C,1-\varepsilon_C],
\end{align*}
$\mu$-almost surely. 
Reversing the calculations above shows that the resulting model reproduces all five moments in $\mathcal D$. 
Hence it is compatible with $\mathcal D$ and satisfies the $\varepsilon$-margin condition.
\end{proof}

\begin{proof}[Proof of Thm.~\ref{thm:margin}]
Suppose that the $\varepsilon$-margin condition is compatible with $\mathcal D$. We proceed in three steps.

First, conditional on a fixed auxiliary treatment stratum $\tilde X=i$, the data determine the marginals $\tilde p_{i1+}$, $\tilde p_{i+0}$, and $\tilde p_{i++}$, while the joint cell
\begin{align*}
    \tilde p_{i10} =
    p(\tilde X=i,
      \tilde Y_{\tilde x}=1,
      \tilde Y_{\tilde x'}=0)
\end{align*}
remains unidentified. The remaining cells in the corresponding $2\times 2$ response table can be written:
\begin{align*}
    \tilde p_{i11}
    &=\tilde p_{i1+}-\tilde p_{i10},\\
    \tilde p_{i00}
    &=\tilde p_{i+0}-\tilde p_{i10},\\
    \tilde p_{i01}
    &=\tilde p_{i++}
      -\tilde p_{i1+}
      -\tilde p_{i+0}
      +\tilde p_{i10}.
\end{align*}
Nonnegativity of these four cells is therefore equivalent to $\tilde\ell_i \leq \tilde p_{i10} \leq \tilde h_i$, where
\begin{align*}
    \tilde\ell_i =
    \max\{
        0,\,
        \tilde p_{i1+}+\tilde p_{i+0}-\tilde p_{i++}
    \}, \quad
    \tilde h_i =
    \min\{
        \tilde p_{i1+},\tilde p_{i+0}
    \}.
\end{align*}
These are precisely the Fr\'echet bounds within auxiliary treatment stratum $i$.

Second, we map the auxiliary response cells back to the original response scale. Since
\begin{align*}
    B = \varepsilon_B+d_B\tilde B, \quad
    1-C = \varepsilon_C+d_C(1-\tilde C),
\end{align*}
we have
\begin{align*}
    B(1-C) ={}
    \varepsilon_B\varepsilon_C
    +\varepsilon_Bd_C(1-\tilde C)
    +d_B\varepsilon_C\tilde B +d_Bd_C\tilde B(1-\tilde C).
\end{align*}
Hence, for each $i\in\{0,1\}$,
\begin{align*}
    \mathbb E[\mathbb I\{\tilde X=i\}B(1-C)] 
    &=
    \varepsilon_B \varepsilon_C \tilde p_{i++}
    + \varepsilon_B d_C \tilde p_{i+0}
    + d_B \varepsilon_C \tilde p_{i1+}
    +d_B d_C \tilde p_{i10}\\
    &=
    f_i(\tilde p_{i10}).
\end{align*}
Finally, since $A=\varepsilon_A+d_A\tilde A$ and $d_A=1-2\varepsilon_A$, it follows that
\begin{align*}
    \mathbb E[AB(1-C)]
    &=
    \varepsilon_A\mathbb E[B(1-C)]
    +d_A\mathbb E[\tilde A B(1-C)]\\
    &=
    \varepsilon_A f_0(\tilde p_{010})
    +(1-\varepsilon_A)f_1(\tilde p_{110}),
\end{align*}
where we used
\begin{align*}
    \mathbb E[B(1-C)]
    &=
    f_0(\tilde p_{010})+f_1(\tilde p_{110}),\\
    \mathbb E[\tilde A B(1-C)]
    &=
    f_1(\tilde p_{110}).
\end{align*}
Similarly,
\begin{align*}
    \mathbb E[(1-A)B(1-C)]
    &=
    (1-\varepsilon_A)f_0(\tilde p_{010})
    +\varepsilon_A f_1(\tilde p_{110}),
\end{align*}
while
\begin{align*}
    \mathbb E[B(1-C)]
    =
    f_0(\tilde p_{010})
    +f_1(\tilde p_{110}).
\end{align*}
Exploiting our linear representation formulae (Appx. \ref{appx:linear}), we have:
\begin{align*}
    \mathrm{PN}
    &=
    \frac{
        \varepsilon_Af_0(\tilde p_{010})
        +(1-\varepsilon_A)f_1(\tilde p_{110})
    }{p(x,y)},\\
    \mathrm{PS}
    &=
    \frac{
        (1-\varepsilon_A)f_0(\tilde p_{010})
        +\varepsilon_Af_1(\tilde p_{110})
    }{p(x',y')},\\
    \mathrm{PNS}
    &=
    f_0(\tilde p_{010})
    +f_1(\tilde p_{110}).
\end{align*}

Because $\varepsilon_J<1/2$ for $J\in\{A,B,C\}$, we have $d_Bd_C>0$, so each $f_i$ is increasing in its argument. 
Moreover, all coefficients multiplying $f_0$ and $f_1$ above are nonnegative.
It follows immediately that all three lower endpoints are obtained by setting
\begin{align*}
    \tilde p_{010}=\tilde\ell_0,
    \qquad
    \tilde p_{110}=\tilde\ell_1,
\end{align*}
and all three upper endpoints by setting
\begin{align*}
    \tilde p_{010}=\tilde h_0,
    \qquad
    \tilde p_{110}=\tilde h_1.
\end{align*}
Substitution yields the stated bounds.

It remains to establish sharpness. For either choice $t_i\in\{\tilde\ell_i,\tilde h_i\}$, define within stratum $i$
\begin{align*}
    \tilde p_{i10}&=t_i,\\
    \tilde p_{i11}&=\tilde p_{i1+}-t_i,\\
    \tilde p_{i00}&=\tilde p_{i+0}-t_i,\\
    \tilde p_{i01}
    &=
    \tilde p_{i++}
    -\tilde p_{i1+}
    -\tilde p_{i+0}
    +t_i.
\end{align*}
By the definitions of $\tilde\ell_i$ and $\tilde h_i$, these cells are nonnegative and preserve all identified auxiliary marginals. The two $\tilde X$ strata may be chosen independently, so taking $(t_0,t_1)=(\tilde\ell_0,\tilde\ell_1)$ or $(t_0,t_1)=(\tilde h_0,\tilde h_1)$ yields a valid auxiliary distribution satisfying the constraints of Prop.~\ref{prop:margin_testability}.

By the converse construction in that proposition, each such auxiliary distribution induces a PNSCM compatible with $\mathcal D$ and satisfying the $\varepsilon$-margin condition. Thus both endpoints of each displayed interval are attainable, proving sharpness.
\end{proof}


\subsection{Soft exogeneity}\label{appx:deviation}

\begin{proof}[Proof of Prop.~\ref{prop:delta_testability}]
Let $V$ denote either $B$ or $C$, and write
\begin{align*}
    m:=\mathbb E[V], \qquad
    q:=\mathbb E[AV].
\end{align*}
Since $\underline a\leq A\leq\overline a$ $\mu$-almost surely,
\begin{align*}
    AV\geq \underline a V,
    \quad
    AV\leq \overline a V,
\end{align*}
and hence
\begin{align*}
    \underline a m\leq q\leq\overline a m.
\end{align*}
Likewise,
\begin{align*}
    A(1-V)\leq\overline a(1-V),
    \quad
    A(1-V)\geq\underline a(1-V).
\end{align*}
Using $\mathbb E[A]=p(x)$ gives
\begin{align*}
    p(x)-\overline a(1-m)
    \leq q
    \leq
    p(x)-\underline a(1-m).
\end{align*}
Combining these inequalities yields
\begin{align*}
    \ell_\delta(m)\leq q\leq h_\delta(m).
\end{align*}
Taking $(m,q)=(p(y_x),p(x,y))$ gives the first condition. For the second, note that
\begin{align*}
    \mathbb E[C]&=p(y_{x'}),\\
    \mathbb E[AC]
    &=p(y_{x'})-\mathbb E[(1-A)C]\\
    &=p(y_{x'})-p(x',y).
\end{align*}
Thus the displayed conditions are necessary.

For sufficiency, suppose first that $\underline a<\overline a$, and set
\begin{align*}
    d:=\overline a-\underline a, \quad
    \alpha:=\frac{p(x)-\underline a}{d}.
\end{align*}
For a pair $(m,q)$ satisfying the displayed inequalities, define
\begin{align*}
    z:=\frac{q-\underline a m}{d}.
\end{align*}
The inequalities $\ell_\delta(m)\leq q\leq h_\delta(m)$ are equivalent to
\begin{align*}
    \max\{0,\alpha+m-1\}
    \leq z
    \leq
    \min\{\alpha,m\}.
\end{align*}
These are precisely the Fr\'echet conditions for a Bernoulli pair $(T,V)$ with
\begin{align*}
    p(T=1)=\alpha,\qquad
    p(V=1)=m,\qquad
    p(T=1,V=1)=z.
\end{align*}
Apply this construction separately to $B$ and $C$, using the same marginal distribution for $T$, and couple $B$ and $C$ conditionally on $T$. Setting $A:=\underline a+dT$ then gives
\begin{align*}
    \mathbb E[A]&=p(x),&
    \mathbb E[B]&=p(y_x),&
    \mathbb E[C]&=p(y_{x'}),\\
    \mathbb E[AB]&=p(x,y),&
    \mathbb E[AC]&=p(y_{x'})-p(x',y),
\end{align*}
while $A\in[\underline a,\overline a]$, $\mu$-almost surely. Equivalently, $\mathbb E[(1-A)C]=p(x',y)$, so the resulting model reproduces $\mathcal D$ and satisfies $\delta$-exogeneity.

If $\underline a=\overline a=p(x)$, corresponding to $\delta=0$, the conditions reduce to $q=p(x)m$ for both pairs, and the same conclusion follows by taking $A\equiv p(x)$. Hence the conditions are sufficient in all cases.
\end{proof}

\begin{proof}[Proof of Thm.~\ref{thm:delta_exogeneity}]
Suppose that $\delta>0$ and that $\delta$-exogeneity is compatible with $\mathcal D$. Recall that
\begin{align*}
    A_\delta
    :=
    \frac{A-\underline a}{d},
    \qquad
    d:=\overline a-\underline a,
\end{align*}
so that
\begin{align*}
    A=\underline a+dA_\delta,
    \qquad
    A_\delta\in[0,1].
\end{align*}
The auxiliary distribution
\begin{align*}
    p^\delta_{ijk}
    :=
    p(X_\delta=i,Y_x=j,Y_{x'}=k)
\end{align*}
therefore satisfies
\begin{align*}
    p^\delta_{ijk}
    =
    \mathbb{E}\!\left[
        A_\delta^i(1-A_\delta)^{1-i}
        B^j(1-B)^{1-j}
        C^k(1-C)^{1-k}
    \right].
\end{align*}

Fix an auxiliary treatment stratum $i\in\{0,1\}$. Its identified marginals determine
\begin{align*}
    p^\delta_{i1+},
    \qquad
    p^\delta_{i+0},
    \qquad
    p^\delta_{i++},
\end{align*}
while the cross-world cell
\begin{align*}
    p^\delta_{i10}
    =
    p(X_\delta=i,Y_x=1,Y_{x'}=0)
\end{align*}
remains unresolved. Writing $t_i:=p^\delta_{i10}$, the remaining three cells in this $2\times2$ response table are
\begin{align*}
    p^\delta_{i11}
    &=p^\delta_{i1+}-t_i,\\
    p^\delta_{i00}
    &=p^\delta_{i+0}-t_i,\\
    p^\delta_{i01}
    &=
    p^\delta_{i++}
    -p^\delta_{i1+}
    -p^\delta_{i+0}
    +t_i.
\end{align*}
Their nonnegativity is equivalent to
\begin{align*}
    \ell_i^\delta
    \leq
    t_i
    \leq
    h_i^\delta,
\end{align*}
where
\begin{align*}
    \ell_i^\delta
    :=
    \max\{
        0,\,
        p^\delta_{i1+}+p^\delta_{i+0}-p^\delta_{i++}
    \}, \quad 
    h_i^\delta
    :=
    \min\{
        p^\delta_{i1+},p^\delta_{i+0}
    \}.
\end{align*}
Thus $\ell_i^\delta$ and $h_i^\delta$ are precisely the Fr\'echet bounds on $p^\delta_{i10}$.

We next express the probabilities of causation in terms of these two free cells. Since
\begin{align*}
    \mathbb E[B(1-C)]
    &=
    p^\delta_{010}+p^\delta_{110},\\
    \mathbb E[A_\delta B(1-C)]
    &=
    p^\delta_{110},
\end{align*}
we obtain
\begin{align*}
    \mathbb E[AB(1-C)]
    &=
    \underline a\,\mathbb E[B(1-C)]
    +d\,\mathbb E[A_\delta B(1-C)]\\
    &=
    \underline a\,p^\delta_{010}
    +(\underline a+d)p^\delta_{110}\\
    &=
    \underline a\,p^\delta_{010}
    +\overline a\,p^\delta_{110}.
\end{align*}
Similarly,
\begin{align*}
    \mathbb E[(1-A)B(1-C)]
    &=
    (1-\underline a)\mathbb E[B(1-C)]
    -d\,\mathbb E[A_\delta B(1-C)]\\
    &=
    (1-\underline a)p^\delta_{010}
    +(1-\overline a)p^\delta_{110},
\end{align*}
while
\begin{align*}
    \mathbb E[B(1-C)]
    =
    p^\delta_{010}+p^\delta_{110}.
\end{align*}
By linearity (Appx. \ref{appx:linear}),
\begin{align*}
    \mathrm{PN}
    &=
    \frac{
        \underline a\,p^\delta_{010}
        +\overline a\,p^\delta_{110}
    }{p(x,y)},\\
    \mathrm{PS}
    &=
    \frac{
        (1-\underline a)p^\delta_{010}
        +(1-\overline a)p^\delta_{110}
    }{p(x',y')},\\
    \mathrm{PNS}
    &=
    p^\delta_{010}+p^\delta_{110}.
\end{align*}

Because
\begin{align*}
    0\leq\underline a\leq\overline a\leq1,
\end{align*}
all coefficients of $p^\delta_{010}$ and $p^\delta_{110}$ in these
expressions are nonnegative. Hence all three lower endpoints are
obtained simultaneously by taking
\begin{align*}
    p^\delta_{010}=\ell^\delta_0,
    \qquad
    p^\delta_{110}=\ell^\delta_1,
\end{align*}
and all three upper endpoints by taking
\begin{align*}
    p^\delta_{010}=h^\delta_0,
    \qquad
    p^\delta_{110}=h^\delta_1.
\end{align*}
Substitution gives the stated bounds.

It remains to establish sharpness. Choose arbitrary $t_i\in[\ell_i^\delta,h_i^\delta]$ for $i\in\{0,1\}$ and define
\begin{align*}
    p^\delta_{i10}&=t_i,\\
    p^\delta_{i11}&=p^\delta_{i1+}-t_i,\\
    p^\delta_{i00}&=p^\delta_{i+0}-t_i,\\
    p^\delta_{i01}
    &=
    p^\delta_{i++}
    -p^\delta_{i1+}
    -p^\delta_{i+0}
    +t_i.
\end{align*}
By construction, these cells are nonnegative and preserve all
identified marginals. The two $X_\delta$ strata can be coupled
independently, so this defines a valid auxiliary distribution
$p^\delta$ for every pair
$(t_0,t_1)$ in the product of the two Fr\'echet intervals.

To recover a compatible PNSCM, introduce one latent state $u_{ijk}$
for every positive cell $p^\delta_{ijk}$, with $p(u_{ijk})=p^\delta_{ijk}$, and set
\begin{align*}
    A_\delta(u_{ijk})&=i,&
    B(u_{ijk})&=j,&
    C(u_{ijk})&=k.
\end{align*}
Finally let
\begin{align*}
    A(u_{ijk})
    :=
    \underline a+d\,A_\delta(u_{ijk}).
\end{align*}
Thus $A$ takes only the values $\underline a$ and $\overline a$, and therefore satisfies $\delta$-exogeneity.

Under compatibility with $\mathcal D$, we can choose $(t_0,t_1)=(\ell^\delta_0,\ell^\delta_1)$ and $(t_0,t_1)=(h^\delta_0,h^\delta_1)$ to attain the stated lower and upper endpoints, respectively. The bounds are therefore sharp.
\end{proof}


\subsection{Stochastic monotonicity}\label{appx:smon}

\begin{proof}[Proof of Prop.~\ref{prop:smono_testability}]
Suppose first that $\mathcal M$ satisfies s-monotonicity, so that
$B\geq C$, $\mu$-almost surely. Since
\begin{align*}
    p(y)
    =
    \mathbb E[AB+(1-A)C],
\end{align*}
we have
\begin{align*}
    p(y_x)-p(y)
    &=
    \mathbb E[B]-\mathbb E[AB+(1-A)C]\\
    &=
    \mathbb E[(1-A)(B-C)]
    \geq0,
\end{align*}
and similarly
\begin{align*}
    p(y)-p(y_{x'})
    &=
    \mathbb E[AB+(1-A)C]-\mathbb E[C]\\
    &=
    \mathbb E[A(B-C)]
    \geq0.
\end{align*}
Thus
\begin{align*}
    p(y_x)\geq p(y)\geq p(y_{x'}).
\end{align*}

Conversely, suppose these inequalities hold. By the cross-world identities Eq. \eqref{eq:crossworld},
\begin{align*}
    p(y\mid x)-p(y_{x'}\mid x)
    &=
    \frac{p(y)-p(y_{x'})}{p(x)}
    \geq0,\\
    p(y_x\mid x')-p(y\mid x')
    &=
    \frac{p(y_x)-p(y)}{p(x')}
    \geq0.
\end{align*}
Define
\begin{align*}
    b_x&:=p(y\mid x),&
    c_x&:=p(y_{x'}\mid x),\\
    b_{x'}&:=p(y_x\mid x'),&
    c_{x'}&:=p(y\mid x').
\end{align*}
Then
\begin{align*}
    0\leq c_x\leq b_x\leq1,
    \qquad
    0\leq c_{x'}\leq b_{x'}\leq1.
\end{align*}

For each $z\in\{x,x'\}$, construct a response distribution supported on
\begin{align*}
    (B,C)\in\{(0,0),(1,0),(1,1)\}
\end{align*}
with respective masses
\begin{align*}
    1-b_z,\qquad b_z-c_z,\qquad c_z.
\end{align*}
Every support point satisfies $B\geq C$, and the resulting means are $\mathbb E[B\mid z]=b_z$ and $\mathbb E[C\mid z]=c_z$.

Finally, assign total mass $p(x)$ to the first response distribution with $A=1$, and total mass $p(x')$ to the second with $A=0$. Then
\begin{align*}
    \mathbb E[A]&=p(x),\\
    \mathbb E[AB]&=p(x)b_x=p(x,y),\\
    \mathbb E[(1-A)C]&=p(x')c_{x'}=p(x',y),
\end{align*}
while
\begin{align*}
    \mathbb E[B]
    &=
    p(x)b_x+p(x')b_{x'}
    =p(y_x),\\
    \mathbb E[C]
    &=
    p(x)c_x+p(x')c_{x'}
    =p(y_{x'}).
\end{align*}
Hence the constructed PNSCM reproduces $\mathcal D$ and satisfies $B\geq C$ $\mu$-almost surely. The condition is therefore sufficient.
\end{proof}

\begin{proof}[Proof of Thm.~\ref{thm:smon}]
By consistency,
\begin{align*}
    p(y_x\mid x)=p(y\mid x), \qquad
    p(y_{x'}\mid x')=p(y\mid x').
\end{align*}
The remaining two stratum-specific counterfactual marginals $p(y_x \mid x')$ and $p(y_{x'} \mid x)$ are identified via Eq. (\ref{eq:crossworld}).

Consider either factual treatment stratum, and temporarily write
\begin{align*}
    b_z=p(y_x\mid z),
    \qquad
    c_z=p(y_{x'}\mid z),
\end{align*}
where $z\in\{x,x'\}$. S-monotonicity holds pointwise after conditioning on either stratum. The sharp moment bounds are therefore
\begin{align}
    b_z-c_z \leq p(y_x,y'_{x'}\mid z) \leq b_z-\frac{c_z^2}{1-b_z+c_z}.
    \label{eq:conditional_smon_bound}
\end{align}
The lower bound follows from
\begin{align*}
    p(y_x,y'_{x'}\mid u) - \big[p(y_x\mid u)-p(y_{x'}\mid u) \big] = p(y_{x'}\mid u)\,p(y'_x\mid u) \geq 0.
\end{align*}
For the upper bound, letting
$b_u=p(y\mid x,u)$ and $c_u=p(y\mid x',u)$ within the stratum,
s-monotonicity implies $b_u(1-b_u+c_u)\geq c_u$. Hence
\begin{align*}
    b_u c_u \geq  \frac{c_u^2}{1-b_u+c_u},
\end{align*}
and Cauchy--Schwarz gives
\begin{align*}
    \mathbb E[b_u c_u\mid z] \geq \frac{c^2}{1-b_z+c_z}.
\end{align*}
Both inequalities are attainable, so
\eqref{eq:conditional_smon_bound} is sharp.

For $\mathrm{PN}$,
\begin{align*}
    \mathrm{PN} = \frac{p(y_x,y'_{x'}\mid x)}{p(y_x\mid x)}.
\end{align*}
Substituting our $p(y_{x'} \mid x)$ formula into
\eqref{eq:conditional_smon_bound}, dividing through by
$p(y\mid x)$, and simplifying gives $\mathrm{PN}_{m} + \psi_{\mathrm{PN}}$.
Similarly,
\begin{align*}
    \mathrm{PS} = \frac{p(y_x,y'_{x'}\mid x')}{p(y'_{x'}\mid x')},
\end{align*}
and the cross-world formula for $p(y_{x} \mid x')$ yields $\mathrm{PS}_{m} + \psi_{\mathrm{PS}}$.
Finally,
\begin{align*}
    \mathrm{PNS} = p(x)\,p(y_x,y'_{x'}\mid x) + p(x')\,p(y_x,y'_{x'}\mid x').
\end{align*}
Applying the two sharp stratum-specific bounds and simplifying yields $\mathrm{PNS}_{m} + \psi_{\mathrm{PNS}}$. Since latent complexity is unrestricted, the two factual-treatment strata can be realized on disjoint latent states, so their extremizers can be attained simultaneously. The resulting PNS bounds are therefore sharp.
\end{proof}

\noindent \textbf{Remark}. We note that, whereas $r_{\mathrm{PN}} := \psi_{\mathrm{PN}} / (1 - \mathrm{PN}_m) \in [0,1]$ and $r_{\mathrm{PS}} := \psi_{\mathrm{PS}} / (1 - \mathrm{PS}_m) \in [0,1]$, the equivalent expression for PNS is somewhat more involved. Write:
\begin{align*}
    w := \frac{p(x, y') + p(x, y_{x'})}{1 - \mathrm{PNS}_{m}}.
\end{align*}
Then $r_{\mathrm{PNS}} := w r_{\mathrm{PN}} (1 - r_{\mathrm{PN}}) + (1 - w) r_{\mathrm{PS}} (1 - r_{\mathrm{PS}})$. As a mixture of two Bernoulli variances, we naturally have $r_{\mathrm{PNS}} \in [0, 1/4]$, so the s-monotone upper bound for $\mathrm{PNS}$ lies at most one quarter of the remaining distance from the classical \citetalias{tian_pearl_2000} value to one. 

\subsection{Personalized treatment decisions}\label{appx:application}

\begin{proof}[Proof of Cor.~\ref{cor:harm}]
By the proof of Prop.~\ref{prop:smono_testability}, the condition $p(y_x) \geq p(y) \geq p(y_{x'})$ is equivalent to $c_z \leq b_z$ for $z \in \{x, x'\}$, where consistency gives 
\begin{align*}
    b_x = p(y \mid x), \quad c_{x'} = p(y \mid x'),
\end{align*}
and $b_{x'}, c_x$ are identifiable via Eq. \eqref{eq:crossworld}.
Since $P(H) = \mathrm{PNS} - \mathrm{ATE}$, it suffices to bound $\mathrm{PNS}$.

For the assumption-free bounds, take $\varepsilon = (0,0,0)$ in Thm.~\ref{thm:margin}. Then $\tilde p_{ijk} = p_{ijk}$ and $f_i(z) = z$, and identifying $i = 1, 0$ with $z = x, x'$, we have 
\begin{align*}
    \tilde p_{i1+} = p(z) b_z, \quad 
    \tilde p_{i+0} = p(z)(1 - c_z), \quad
    \tilde p_{i++} = p(z).
\end{align*}
The sharp bounds of Thm.~\ref{thm:margin} therefore become
\begin{align*}
    \sum_z p(z) \max\{0,\, b_z - c_z\} \leq \mathrm{PNS} \leq \sum_z p(z) \min\{b_z,\, 1 - c_z\},
\end{align*}
i.e.\ the stratum-wise Fr\'echet bounds averaged over $z$. Since $b_z \geq c_z$, the lower bound equals 
\begin{align*}
    \sum_z p(z)(b_z - c_z) = \mathrm{ATE}.
\end{align*}
Subtracting $\mathrm{ATE}$ and using 
\begin{align*}
    \min\{b_z, 1-c_z\} - b_z + c_z = \min\{c_z, 1 - b_z\}
\end{align*}
gives the sharp bounds $0 \leq P(H) \leq \tau$.

Under s-monotonicity, Thm.~\ref{thm:smon} gives $\mathrm{PNS}_m = \mathrm{ATE}$, and subtracting $b_z - c_z$ from the upper bound in \eqref{eq:conditional_smon_bound} gives the stratum-specific bound
\begin{align*}
    p(y'_x, y_{x'} \mid z) \leq c_z - \frac{c_z^2}{1 - b_z + c_z} = \frac{c_z (1 - b_z)}{c_z + 1 - b_z},
\end{align*}
whose average over $z$ is $\psi_{\mathrm{PNS}}$. 

For the final claim, let $m_z := \min\{c_z, 1 - b_z\}$ and $M_z := \max\{c_z, 1 - b_z\}$. Then the stratum term of $\psi_{\mathrm{PNS}}$ equals 
\begin{align*}
    \frac{m_z M_z}{m_z + M_z},
\end{align*}
(or zero if $m_z + M_z = 0$), which lies in $[m_z/2, m_z]$ because $m_z \leq M_z$. 
Averaging over $z$ gives $\tau/2 \leq \psi_{\mathrm{PNS}} \leq \tau$. 
Further, writing $t_z := c_z / (1 - b_z + c_z)$, the stratum term equals 
\begin{align*}
    (1 - (b_z - c_z))\, t_z (1 - t_z) \leq \frac{(1 - (b_z - c_z))}{4},
\end{align*}
and averaging over $z$ gives $\psi_{\mathrm{PNS}} \leq (1 - \mathrm{ATE})/4$.
Finally, $(1-\mathrm{ATE})/4 < \mathrm{ATE}/\lambda$ iff $\mathrm{ATE} > \lambda/(\lambda + 4)$.
\end{proof}

\paragraph{The default action.} 
The probabilities of benefit and harm, and hence rule \eqref{eq:rule}, presuppose a distinction between a default action and a deviant one. \citet{mueller2023} leave this implicit, whereas \citet{richens2022} make it explicit: harm is defined relative to a default action, here $x'$. 
In a medical context, the default is the existing baseline (no treatment, or the standard of care), and the deviant action is a novel intervention; \citet{sarvet_stensrud_2025} describe the same asymmetry as one between errors of commission and errors of omission. 
Rule \eqref{eq:rule} expresses the idea that a deviant action should meet a higher standard than the default. 
Our analysis takes this premise as given, and shows what the rule requires once determinism is dropped; for a critical discussion of the premise and of the resulting notion of harm, see \citet{beckers2024causal, beckers2023harm, beckers2026harm}.

\paragraph{Harm without determinism.} 
In a deterministic model, every individual has a fixed response type, and $P(H)$ is the proportion of individuals whom treatment would harm. 
In a PNSCM, $u$ fixes only the response probabilities, and $P(H \mid u) = p(y' \mid x, u)\, p(y \mid x', u)$ is the probability of an event: the deviant action is taken, the outcome is bad, and under the default it would have been good. 
This probability is positive for almost every latent state, including those at which treatment increases the probability of recovery. 
The rationale for rule \eqref{eq:rule} is unaffected, since it concerns precisely such events rather than the existence of a harmed type. 
What changes is the calibration of $\lambda$: in a deterministic model, a safe treatment has $P(H) = 0$, whereas in a PNSCM, almost no treatment does, so a value of $\lambda$ chosen with the former in mind will be more demanding than intended.

\paragraph{Further details on the example.} 
The upper bound $\psi_{\mathrm{PNS}}$ is attained by a model in which, within stratum $z$, a fraction $b_z - c_z$ recover iff treated, while the remaining fraction $1 - b_z + c_z$ recover with probability $c_z/(1 - b_z + c_z)$ regardless of treatment, in the sense that $B(u) = C(u) = c_z/(1 - b_z + c_z)$ while $Y_x$ and $Y_{x'}$ are drawn independently given $u$; it is precisely in this subpopulation that harm occurs, with probability $(1 - B(u))\,C(u) > 0$ conditional on $u$. In the example, this means that $30\%$ of the treated recover iff treated while the other $70\%$ recover with probability $1/7$; meanwhile, $23\%$ of the untreated recover iff treated while the other $77\%$ recover with probability $14/23$. 
For comparison, consider the data of \citet[Example~4]{dawid_senn_2025}, originally due to \citet{mueller2023}: 
\begin{equation*}
\renewcommand{\arraystretch}{1.6}
\begin{array}{r@{\;}c@{\;}l | r@{\;}c@{\;}l}
\multicolumn{3}{c|}{\textbf{obs}} & \multicolumn{3}{c}{\textbf{int}} \\[2pt]
\hline
p(x, y)   &=& 0.2  & p(y_x)    &=& 0.49 \\
p(x, y')  &=& 0.5  & p(y_{x'}) &=& 0.21 \\
p(x', y)  &=& 0.1  & \multicolumn{3}{c}{} \\
p(x', y') &=& 0.2  & \multicolumn{3}{c}{} \\
\end{array}
\end{equation*}
Then 
\begin{align*}
    (b_x, c_x, b_{x'}, c_{x'}) = (0.286, 0.157, 0.967, 0.333),
\end{align*}
so that 
\begin{align*}
    \tau = 0.7 \cdot 0.157 + 0.3 \cdot 0.033 = 0.12 < \mathrm{ATE}/2 = 0.14.
\end{align*}
Hence rule \eqref{eq:rule} with $\lambda = 2$ already justifies treatment under the assumption-free bounds ($\lambda^* = 2.33$), even though $P(H) \in [0, 0.12]$ is not point identified, and no resolution of the remaining ambiguity, such as the midpoints considered by \citet{dawid_senn_2025}, is required.
Here $\psi_{\mathrm{PNS}} = 0.099$, so s-monotonicity raises $\lambda^*$ to $2.82$, and it therefore only affects the decision for $\lambda \in (2.33, 2.82]$. 
For this reason the main text uses observational data for which the assumption-free bounds do not suffice at $\lambda = 2$.

\section{Simulations}\label{appx:experiments}

In this section, we present results from a supplemental simulation experiment sweeping a grid of hyperparameter values.
We assess the sensitivity of our identification results across 18 five-state PNSCMs, comprising nine common-margin designs (M11--M33) and nine soft-exogeneity designs (D11--D33). 
Each population is specified by latent masses $w_u=p(u)$, treatment propensities $a_u=p(x\mid u)$, and response kernels $b_u=p(y_x\mid u)$ and $c_u=p(y_{x'}\mid u)$, with $p(x)=1/2$. 
To obtain a manageable but heterogeneous grid, we use a three-level, strength-two orthogonal-array design over four primary targets: 
the oracle margin $\varepsilon^*\in\{0.035,0.070,0.110\}$, the oracle exogeneity tolerance $\delta^*\in\{0.070,0.150,0.300\}$, the normalized absolute confounding bias $|\mathrm{RD}_{\mathrm{obs}}-\mathrm{ATE}|/\delta^* \in\{0.2,0.5,0.8\}$, and $\mathrm{ATE}\in\{-0.12,0.05,0.18\}$.
We additionally target varying distances between the oracle restriction and the boundary of data compatibility: $\varepsilon_{\max}-\varepsilon^* \in\{0.05,0.08,0.12\}$ in the margin designs, and $\delta_{\min}/\delta^*\in\{0.30,0.60,0.85\}$ in the exogeneity designs. 
These secondary targets are not independently crossed with all four primary factors. 
Models are selected to approximately match these targets from candidate latent mechanisms. 
The oracle restriction levels are exact, whereas the compatibility, confounding, and ATE targets need not all be jointly attainable. 

\paragraph{Results.}
The appendix grids display sharp bounds on PN, PS, and PNS as $\varepsilon$ increases or $\delta$ decreases, against the unrestricted \citetalias{tian_pearl_2000} intervals. 
Grey shading denotes the assumption-free bounds, blue shading our restricted sharp bounds, and the solid black line the model-implied true probability of causation. 
As in Fig.~\ref{fig:semi_synthetic}, the dotted black line indicates the oracle restriction, while the red line marks the maximum data-compatible margin $\varepsilon_{\max}$ or minimum data-compatible tolerance $\delta_{\min}$.
For soft exogeneity, dashed green lines additionally show the classical exact-exogeneity bounds, which are not generally valid under the actual data-generating model.

At the oracle values, the nine margin-targeted populations yield median width reductions of $44.4\%$, $46.1\%$, and $43.7\%$ for PN, PS, and PNS, respectively; the nine exogeneity-targeted populations yield $24.3\%$, $32.1\%$, and $28.2\%$.
PNS gains range from $17.6\%$ to $68.4\%$ under the common margin and from $15.5\%$ to $60.5\%$ under soft exogeneity.
All oracle-calibrated intervals contain their respective model-implied ground truths.
The variation in gains illustrates that neither restriction is uniformly informative. 
Contributions depend on both permissible latent heterogeneity and the observational--interventional discrepancies they must explain.
We highlight that some values of $\varepsilon, \delta$ may be compatible with $\mathcal D$ but over-optimistic, leading to invalid bounds. 
This is not uncommon when partial identification intervals depend on latent parameters; see, e.g., \citep[Thm.~1]{watson2024bounding}.

\begin{sidewaysfigure}
    \centering
    \includegraphics[width=0.95\textheight, keepaspectratio]{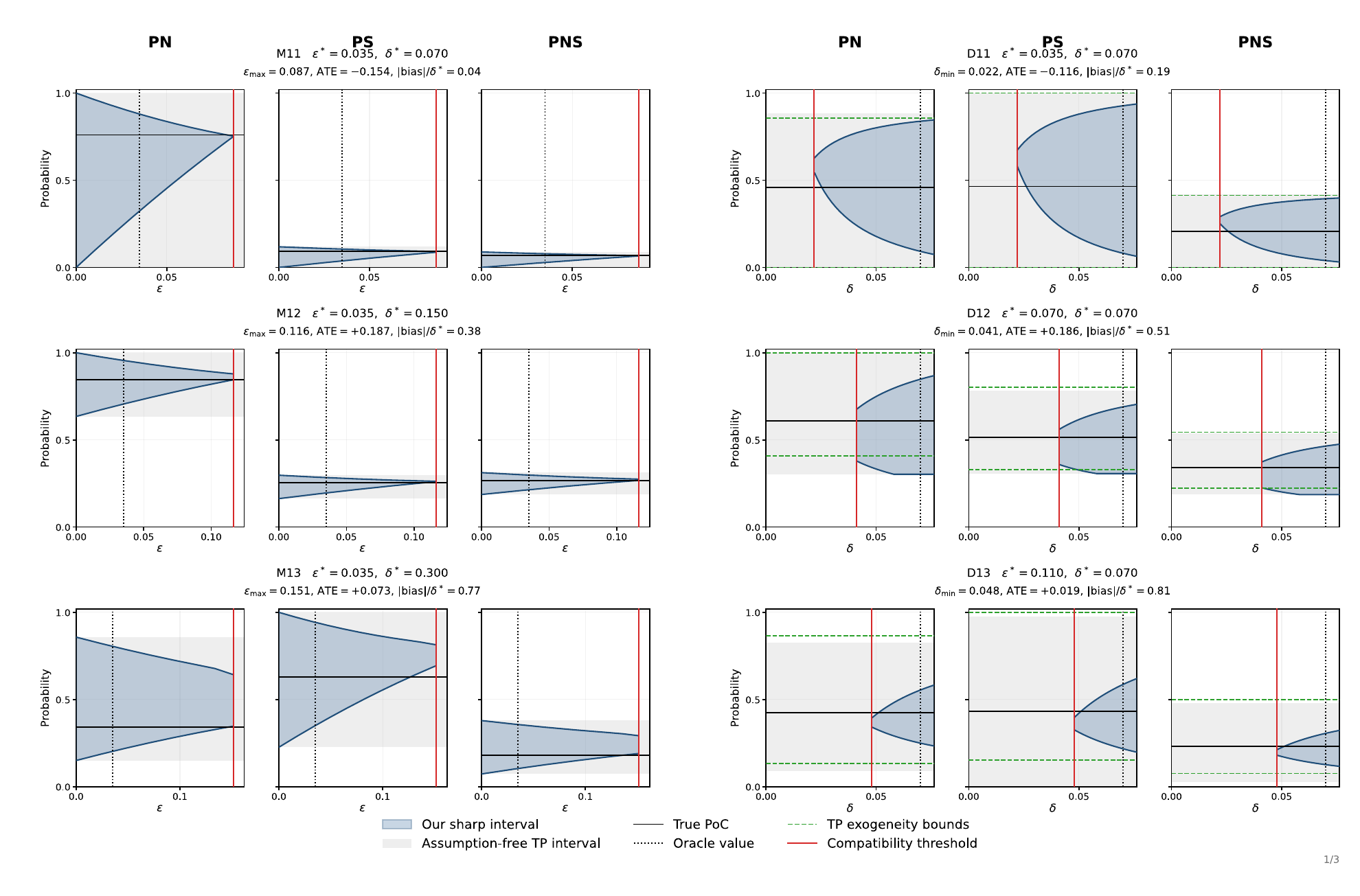}
    \label{fig:page1}
\end{sidewaysfigure}

\begin{sidewaysfigure}
    \centering
    \includegraphics[width=0.95\textheight, keepaspectratio]{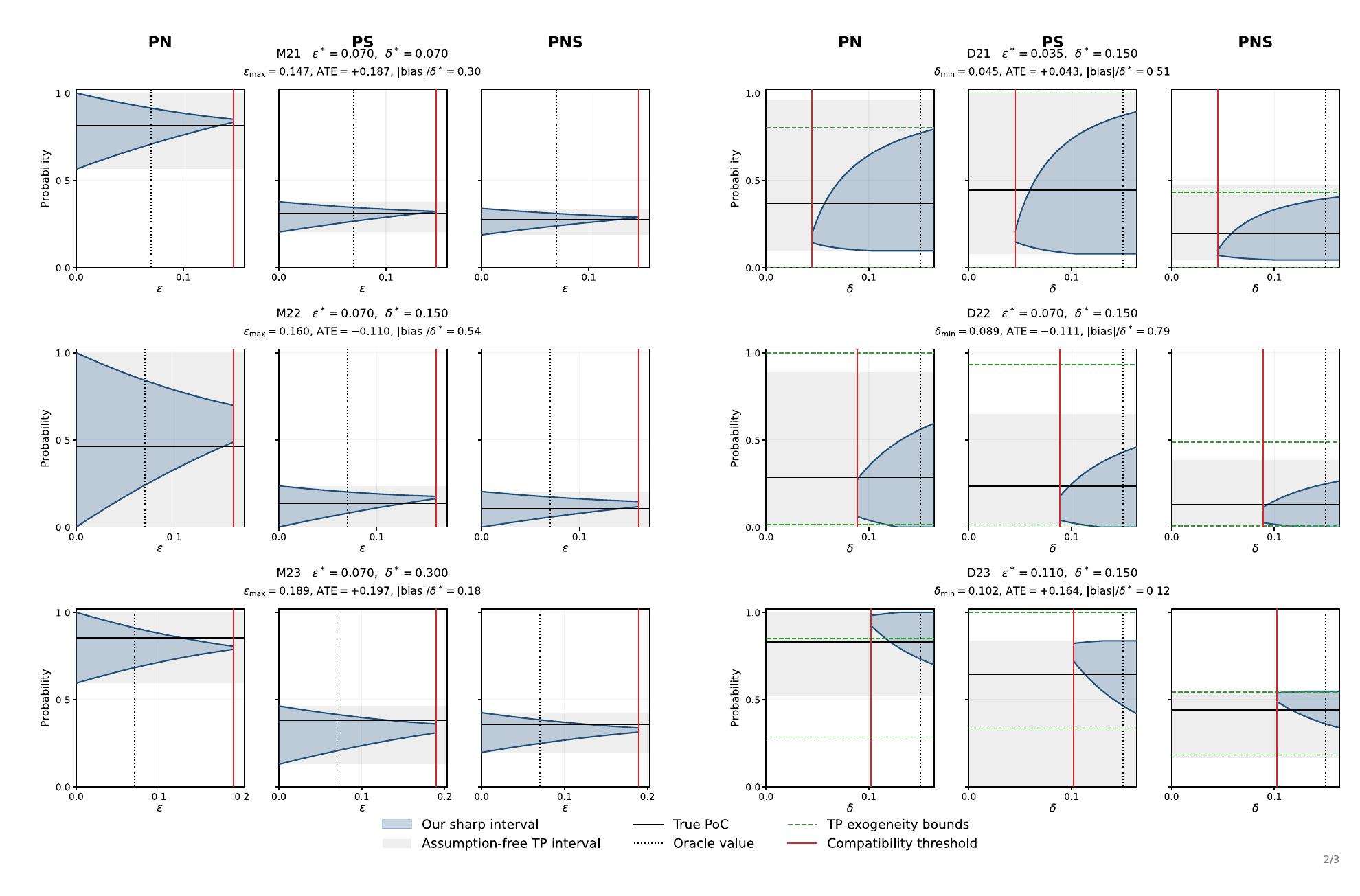}
    \label{fig:page2}
\end{sidewaysfigure}

\begin{sidewaysfigure}
    \centering
    \includegraphics[width=0.95\textheight, keepaspectratio]{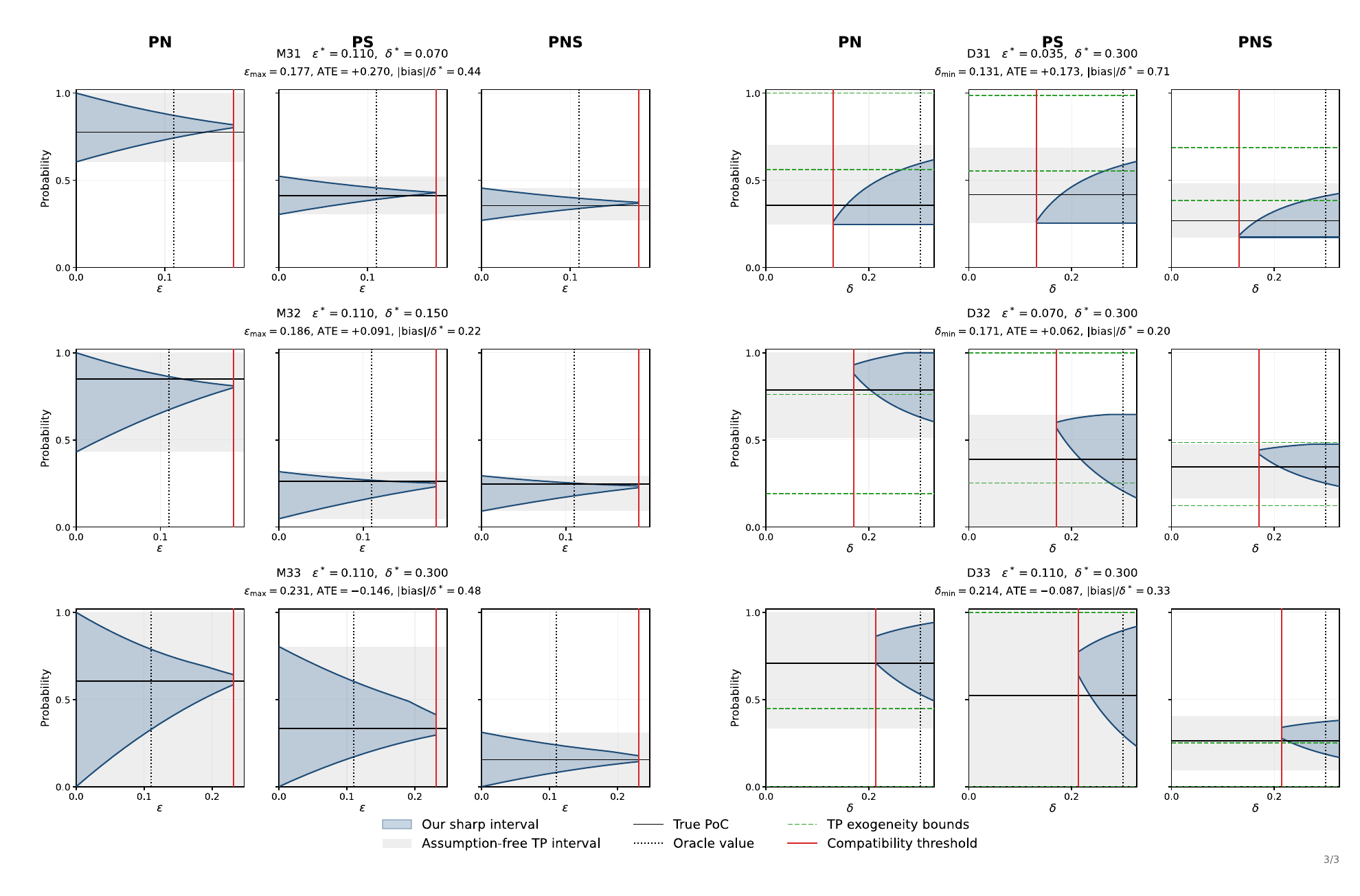}
    \label{fig:page3}
\end{sidewaysfigure}

%% file: biblio.bib
@book{pearl2009causality,
  title={Causality},
  address = {New York},
  author={Pearl, Judea},
  year={2009},
  publisher={Cambridge University Press}
}

@inproceedings{beckers2025nondeterministic,
title={Nondeterministic Causal Models},
author={Sander Beckers},
booktitle={Fourth Conference on Causal Learning and Reasoning},
year={2025}
}

@misc{beckers2025b,
  author = {Sander Beckers},
  title  = {Causal Counterfactuals Reconsidered},
  year   = {2025},
  note   = {arxiv.org/pdf/2512.12804}
}

@article{tian_pearl_2000,
	author = {Tian, Jin and Pearl, Judea},
	journal = {Annals of Mathematics and Artificial Intelligence},
	number = {1},
	pages = {287--313},
	title = {Probabilities of causation: Bounds and identification},
	volume = {28},
	year = {2000}
}

@article{shpitser2008,
  author  = {Ilya Shpitser and Judea Pearl},
  title   = {Complete Identification Methods for the Causal Hierarchy},
  journal = {Journal of Machine Learning Research},
  year    = {2008},
  volume  = {9},
  number  = {64},
  pages   = {1941--1979}
}

@article{balke97,
	author = {Alexander Balke and Judea Pearl},
	journal = {Journal of the American Statistical Association},
	number = {439},
	pages = {1171--1176},
	title = {Bounds on Treatment Effects from Studies with Imperfect Compliance},
	volume = {92},
	year = {1997}
}

@inproceedings{jiang2023,
author = {Jiang, Ziwei and Wei, Lai and Kocaoglu, Murat},
title = {Approximate causal effect identification under weak confounding},
year = {2023},
booktitle = {Proceedings of the 40th International Conference on Machine Learning}
}

@article{watson2022,
	author = {Watson, David S. and Gultchin, Limor and Taly, Ankur and Floridi, Luciano},
	journal = {Minds and Machines},
	number = {1},
	pages = {185--218},
	title = {Local Explanations via Necessity and Sufficiency: Unifying Theory and Practice},
	volume = {32},
	year = {2022}
}

@article{Gupta2011IntentionToTreat,
  author  = {Gupta, Sandeep K.},
  title   = {Intention-to-treat concept: A review},
  journal = {Perspectives in Clinical Research},
  year    = {2011},
  volume  = {2},
  number  = {3},
  pages   = {109--112}
}

@article{halpern_pearl,
	author = {Halpern, Joseph Y. and Pearl, Judea},
	isbn = {0007-0882},
	journal = {The British Journal for the Philosophy of Science},
	number = {4},
	pages = {889--911},
	title = {Causes and Explanations: A Structural-Model Approach. {Part II}: Explanations},
	volume = {56},
	year = {2005}
}

@inbook{bareinboim2022,
author = {Bareinboim, Elias and Correa, Juan D. and Ibeling, Duligur and Icard, Thomas},
title = {On Pearl’s Hierarchy and the Foundations of Causal Inference},
year = {2022},
booktitle = {Probabilistic and Causal Inference: The Works of Judea Pearl},
pages = {507–556},
}

@article{Robins1989ThePO,
  title={The probability of causation under a stochastic model for individual risk.},
  author={James M. Robins and Sander Greenland},
  journal={Biometrics},
  year={1989},
  volume={45 4},
  pages={1125-1138}
}

@article{pearl99,
	author = {Pearl, Judea},
	journal = {Synthese},
	number = {1},
	pages = {93--149},
	title = {Probabilities Of Causation: Three Counterfactual Interpretations And Their Identification},
	volume = {121},
	year = {1999}
}

@inproceedings{balke1994,
	author = {Alexander Balke and Judea Pearl},
	booktitle = {Uncertainty in Artificial Intelligence},
	pages = {46-54},
	title = {Counterfactual Probabilities: Computational Methods, Bounds and Applications},
	year = {1994}
}

@book{lewis1973,
  author    = {Lewis, David},
  title     = {Counterfactuals},
  year      = {1973},
  publisher = {Blackwell},
  address   = {Oxford}
}

@article{dawid2000,
  author  = {Dawid, A. Philip},
  title   = {Causal Inference without Counterfactuals},
  journal = {Journal of the American Statistical Association},
  year    = {2000},
  volume  = {95},
  number  = {450},
  pages   = {407--424},
  doi     = {10.1080/01621459.2000.10474210}
}

@inproceedings{galhotra_halpern_2024,
  author    = {Galhotra, Sainyam and Halpern, Joseph Y.},
  title     = {Intervention and Conditioning in Causal {B}ayesian Networks},
  booktitle = {Advances in Neural Information Processing Systems},
  year      = {2024},
  volume    = {37},
  pages     = {89019--89041}
}

@article{rubin1974,
  author  = {Rubin, Donald B.},
  title   = {Estimating Causal Effects of Treatments in Randomized and Nonrandomized Studies},
  journal = {Journal of Educational Psychology},
  year    = {1974},
  volume  = {66},
  number  = {5},
  pages   = {688--701}
}

@InProceedings{beckers2022,
  title = 	 {Causal Explanations and {XAI}},
  author =       {Beckers, Sander},
  booktitle = 	 {Proceedings of the First Conference on Causal Learning and Reasoning},
  pages = 	 {90--109},
  year = 	 {2022}
}

@article{verma2024,
author = {Verma, Sahil and Boonsanong, Varich and Hoang, Minh and Hines, Keegan and Dickerson, John and Shah, Chirag},
title = {Counterfactual Explanations and Algorithmic Recourses for Machine Learning: A Review},
year = {2024},
volume = {56},
number = {12},
journal = {ACM Comput. Surv.}
}

@inproceedings{karimi2021,
author = {Karimi, Amir-Hossein and Sch{\"o}lkopf, Bernhard and Valera, Isabel},
title = {Algorithmic Recourse: from Counterfactual Explanations to Interventions},
year = {2021},
booktitle = {Proceedings of the 2021 ACM Conference on Fairness, Accountability, and Transparency},
pages = {353–362}
}

@article{ramsahai2012,
  author  = {Roland R. Ramsahai},
  title   = {Causal Bounds and Observable Constraints for Non-deterministic Models},
  journal = {J. Mach. Learn. Res.},
  year    = {2012},
  volume  = {13},
  number  = {29},
  pages   = {829--848}
}

@book{Imbens:2015fk,
	Author = {G.~W. Imbens and D.~B. Rubin},
	Publisher = {Cambridge University Press},
	Title = {Causal Inference in Statistics, Social, and Biomedical Sciences: {A}n Introduction},
	Year = {2015}
}

@article{silva_evans:16,
 author = {Ricardo Silva and Robin Evans},
 title = {Causal inference through a witness protection program},
 journal = {Journal of Machine Learning Research},
 volume = {17},
 number = {56},
 pages = {1--53},
 year = {2016}
}

@book{hernan2025causal,
  author    = {Hernan, Miguel A. and Robins, James M.},
  title     = {Causal Inference: What If},
  publisher = {CRC Press},
  address   = {Boca Raton},
  year      = {2025}
}

@article{rosenbaum1983central,
  author  = {Rosenbaum, Paul R. and Rubin, Donald B.},
  title   = {The central role of the propensity score in observational studies for causal effects},
  journal = {Biometrika},
  year    = {1983},
  volume  = {70},
  number  = {1},
  pages   = {41--55}
}

@inproceedings{watson2024bounding,
      title={Bounding Causal Effects with Leaky Instruments}, 
      author={David S. Watson and Jordan Penn and Lee M. Gunderson and Gecia Bravo-Hermsdorff and Afsaneh Mastouri and Ricardo Silva},
      year={2024},
      booktitle={Proceedings of the 40th Conference on Uncertainty in Artificial Intelligence}
}

@book{woodward2003making,
  title     = {Making Things Happen: A Theory of Causal Explanation},
  author    = {Woodward, James},
  year      = {2003},
  publisher = {Oxford University Press},
  address   = {Oxford}
}

@article{Conley2012Plausibly,
  title     = {Plausibly Exogenous},
  author    = {Conley, Timothy G. and Hansen, Christian B. and Rossi, Peter E.},
  journal   = {The Review of Economics and Statistics},
  volume    = {94},
  number    = {1},
  pages     = {260--272},
  year      = {2012}
}

@article{mueller2023,
  author  = {Mueller, Scott and Pearl, Judea},
  title   = {Personalized Decision Making -- A Conceptual Introduction},
  journal = {Journal of Causal Inference},
  volume  = {11},
  number  = {1},
  pages   = {20220050},
  year    = {2023}
}

@article{dawid_senn_2025,
  author  = {Dawid, A. Philip and Senn, Stephen},
  title   = {Personalised Decision-Making without Counterfactuals},
  journal = {Observational Studies},
  note    = {To appear. arXiv:2301.11976},
  year    = {2025}
}

@article{sarvet_stensrud_2025,
  author  = {Sarvet, Aaron L. and Stensrud, Mats J.},
  title   = {Perspectives on ``Harm'' in Personalized Medicine},
  journal = {American Journal of Epidemiology},
  volume  = {194},
  number  = {6},
  pages   = {1743--1748},
  year    = {2025}
}

@misc{sarvet_stensrud_2024,
  author = {Sarvet, Aaron L. and Stensrud, Mats J.},
  title  = {Rejoinder to ``Perspectives on `Harm' in Personalized Medicine -- An Alternative Perspective''},
  year   = {2024},
  note   = {arXiv:2403.14869}
}

@inproceedings{richens2022,
  author    = {Richens, Jonathan G. and Beard, Rory and Thompson, Daniel H.},
  title     = {Counterfactual Harm},
  booktitle = {Advances in Neural Information Processing Systems},
  volume    = {35},
  year      = {2022}
}

@misc{beckers2026harm,
  author = {Beckers, Sander and Chockler, Hana and Halpern, Joseph Y.},
  title  = {Quantifying Harm (extended version)},
  year   = {2026},
  note   = {arXiv:2209.15111}
}

@inproceedings{beckers2023harm,
  author    = {Beckers, Sander and Chockler, Hana and Halpern, Joseph Y.},
  title     = {Quantifying Harm},
  booktitle = {Proceedings of the Thirty-Second International Joint Conference on Artificial Intelligence ({IJCAI}-23)},
  pages     = {363--371},
  year      = {2023},
  doi       = {10.24963/ijcai.2023/41}
}

@article{beckers2024causal,
  author  = {Beckers, Sander and Chockler, Hana and Halpern, Joseph Y.},
  title   = {A Causal Analysis of Harm},
  journal = {Minds and Machines},
  volume  = {34},
  pages   = {34},
  year    = {2024},
  doi     = {10.1007/s11023-024-09689-7}
}

@misc{becker1996adult,
  author       = {Becker, Barry and Kohavi, Ronny},
  title        = {Adult},
  year         = {1996},
  howpublished = {UCI Machine Learning Repository},
  url          = {https://archive.ics.uci.edu/dataset/2/adult}
}

@article{vinokur1995impact,
  author  = {Vinokur, Amiram D. and Price, Richard H. and Schul, Yaacov},
  title   = {Impact of the JOBS intervention on unemployed workers varying in risk for depression},
  journal = {American Journal of Community Psychology},
  year    = {1995},
  volume  = {23},
  number  = {1},
  pages   = {39--74}
}

@inproceedings{lally2026robust,
  author    = {Jessica Lally and Milad Kazemi and Nicola Paoletti and David Watson and Sander Beckers},
  title     = {Robust Counterfactual Policy Optimisation via Nondeterministic Causal Models},
  booktitle = {UAI 2026 Workshop on Causality for Decision Making},
  year      = {2026}
}
